\documentclass{article}

\usepackage{arxiv}

\usepackage[utf8]{inputenc} 
\usepackage[T1]{fontenc}    

\usepackage{url}            
\usepackage{booktabs}       
\usepackage{amsfonts}       
\usepackage{nicefrac}       
\usepackage{microtype}      
\usepackage{lipsum}
\usepackage{graphicx}
\graphicspath{ {./images/} }

\usepackage[utf8]{inputenc}
\usepackage[utf8]{inputenc}
\usepackage{rotating}
\usepackage{tikz}
\usepackage[T1]{fontenc}
\usepackage{amsmath}
\usepackage{graphicx}
\usepackage{svg}
\usepackage{tikz}
\usepackage{lmodern}
\usepackage{verbatim}
\usepackage{graphicx}
\usepackage[figurename=Fig.,labelfont=bf,labelsep=period]{caption}
\usepackage{subcaption}
\usepackage{amsmath}
\usepackage{newtxtext,newtxmath}
\usepackage{tikz}
\usepackage{tabularx}
\usepackage{pgfplots}
\usetikzlibrary{arrows.meta, shapes.multipart, shadows, backgrounds}
\pgfplotsset{compat=1.18}
\usepackage[colorlinks=true,citecolor=red,linkcolor=black]{hyperref}

\title{VectorGraphNET: Graph Attention Networks for Accurate Segmentation of Complex Technical Drawings}

\author{
 Andrea Carrara, Stavros Nousias, André Borrmann\\
  Technical University of Munich\\
  School of Engineering and Design\\
  Chair of Computational Modelling and Simulation\\
    \texttt{andrea.carrara@tum.de}
}

\begin{document}
\maketitle
\begin{abstract}
This paper introduces a new approach to extract and analyze vector data from technical drawings in PDF format. 
Our method involves converting PDF files into SVG format and creating a feature-rich graph representation, which captures the relationships between vector entities using geometrical information. We then apply a graph attention transformer with hierarchical label definition to achieve accurate line-level segmentation. Our approach is evaluated on two datasets, including the public FloorplanCAD dataset, which achieves state-of-the-art results on weighted F1 score, surpassing existing methods. 
The proposed vector-based method offers a more scalable solution for large-scale technical drawing analysis compared to vision-based approaches, while also requiring significantly less GPU power than current state-of-the-art vector-based techniques. Moreover, it demonstrates improved performance in terms of the weighted F1 (wF1) score on the semantic segmentation task.
Our results demonstrate the effectiveness of our approach in extracting meaningful information from technical drawings, enabling new applications, and improving existing workflows in the AEC industry. Potential applications of our approach include automated building information modeling (BIM) and construction planning, which could significantly impact the efficiency and productivity of the industry.
\end{abstract}

\section{Introduction}
\label{sec:Introduction}
The Architecture, Engineering, and Construction (AEC) industry relies heavily on technical drawings to convey complex design information and facilitate stakeholder collaboration. 
However, the traditional manual review and analysis of technical drawings can be time-consuming, prone to human error, and limited in scope \cite{borrmann2018building}. The digitization of technical drawings and the application of automatic content analysis can unlock new insights, enhance project outcomes, and drive innovation in the AEC sector. \cite{moreno2019new}

Despite the growing importance of Building Information Modeling (BIM), traditional technical drawings remain the primary mode of communication between architects/engineers and constructors and are often the sole source of geometric and semantic information for many existing buildings. The development of advanced technologies, such as computer vision and machine learning, has enabled the automation of technical drawing analysis, unlocking new opportunities for data-driven decision-making and improved project delivery. Additionally, the analysis of technical drawings by human experts hinders the optimization of time and cost for an estimated 65\% of companies. In contrast, adopting digitalization and Building Information Models has been shown to significantly reduce documentation time by 75\% \cite{9080390}.

In the construction industry, digital documents are currently the prevailing standard for sharing plans and documents among project stakeholders \cite{RASMUSSEN2019102956}. Specifically, PDF is currently one of the dominant formats for sharing plans and documents. Its widespread adoption is due to its ability to preserve document formatting and layout, ensuring that complex designs and technical specifications are accurately conveyed. However, this reliance on PDF documents also introduces several challenges. For instance, PDFs can be cumbersome to work with, making it difficult to extract and manipulate data, leading to inefficiencies and errors. Moreover, the lack of standardized metadata and inconsistent naming conventions can make searching, organizing, and managing extensive collections of PDF files challenging.
Additionally, the static nature of PDFs can hinder collaboration and real-time feedback, as changes and revisions can be challenging to track and implement. As a result, construction projects often struggle with version control, data inconsistencies, and communication breakdowns, ultimately affecting project timelines, budgets, and overall quality \cite{tariq2023study}. Conversion of PDF-based technical drawings to BIM models can be laborious, expensive, and slow, requiring specialized expertise. Automating this process could significantly accelerate BIM adoption and enhance digitization efforts in the AEC industry.  

This study is building upon preliminary findings described in \cite{carrara2024employing} and explores the application of modern AI techniques to make technical drawing content computationally accessible, leveraging standardized PDF formats to extract vector data and reconstruct semantic information at the line level. Approaches based on pixel images and the application of convolutional neural networks (CNN) have been presented in the past. However, they face significant limitations when applied to more complex drawings with many overlapping lines, as often found in real-world documents. In this study, we show that by employing graph neural networks (GNN) to learn the relationship between vectors and recover semantic information from drawings, the analysis of AEC technical drawings can be dramatically improved.

We build upon the resolution-agnostic vector format of technical drawings, demonstrating a prominent level of precision on the line elements and facilitating the direct use of these vector representations for technical drawing analysis and semantic segmentation. Specifically, our approach extracts the geometrical data for each component and transforms it into graph features. The drawings' vectors are connected using spatially inspired rules that compute geometrical relationships for graph edge feature-based descriptors. The spatial vector descriptor prepares the data for utilization within the GNN architecture. Specifically, we design and test a novel graph attention-based network architecture comprising two layers of graph attention transformer layers, which receive a feature-rich graph representation as input and utilize the message-passing mechanism in graphs to perform semantic segmentation.

The rest of the paper is organized as follows. The RELATED WORK section discusses the existing literature on segmentation in technical drawings, highlighting the key findings, methodologies, and limitations of previous studies. The PRELIMINARIES section provides an overview of the critical concepts of Scalable Vector Graphics (SVG), which is the primary format used for drawings in this study. The METHOD section describes the research design, graph creation algorithm, and analysis procedures, as well as the classification algorithms used in the study. The EXPERIMENTAL SETUP section presents the datasets used in the case study and introduces the definition of the hierarchical labels. The RESULTS section presents the study's findings, including the results of experiments conducted on the two different datasets. The DISCUSSION section interprets the results, highlighting the implications, strengths, and limitations of the study and comparing the findings with existing research. Finally, the CONCLUSION section summarizes the study's main contributions, reiterates the key findings, and outlines potential avenues for future research.

\section{Related work}
\label{sec:related_work}
The automated analysis of technical drawings is a complex task due to the inherent intricacies of the data. The presence of overlapping elements and symbols, such as text, measurement lines, and design features, contributes to this complexity. Furthermore, the variability in the appearance of these elements within the drawings adds to the challenge. Initial research efforts in this domain employed a combination of rule-based analysis and traditional computer vision techniques to address these issues (\cite{Mac2010AST}; \cite{Ahmed2011ImprovedAA}).
Subsequent studies have primarily approached this problem as a computer vision task, focusing on vectorization and semantic segmentation \cite{kalervo2019cubicasa5kdatasetimprovedmultitask}; \cite{8237503}; \cite{7986875}. Vectorization, in this context, refers to the process of converting a raster image, composed of pixels, into a vector representation, which is a set of geometric primitives such as lines, curves, and shapes.
An increasing trend is using deep learning for semantic segmentation of pixel images to address floorplans' high diversity and variability. In this context, various deep learning methods have been explored, including Convolutional Neural Networks (CNNs) \cite{10.1145/3441250.3441265},\cite{9577792},\cite{9009528}, \cite{10.1007/978-981-15-6648-6_2}, Generative Adversarial Networks (GANs) \cite{zhang2020directionawarelearnableadditivekernels}, \cite{info12050206}, and Fully Convolutional Networks (FCNs) \cite{10.1145/3210499.3210528}, \cite{10.1145/3638584.3638636}. These approaches have enabled us to move beyond heuristic-based assumptions and develop more robust and accurate models.
Recently, there has been a shift towards utilizing graph neural networks (GNNs) for element detection and segmentation in technical drawings.

Notably, \cite{9506514} proposed a method for extracting floor plan graphs from CAD primitives and employed GNNs to classify graph nodes, thereby differentiating between doors and non-doors. This approach facilitated general room detection in office buildings. It demonstrated improved performance compared to earlier techniques, such as those presented by \cite{7986875} and the Faster R-CNN baseline \cite{7410526}.

\cite{fan2021floorplancadlargescalecaddrawing} have also explored the application of GNNs for panoptic segmentation, aiming to detect countable objects like furniture, windows, and doors in CAD drawings and establish the semantics of uncountable elements, such as walls. Their work involves a novel algorithm that combines GNNs with CNNs for symbol spotting within the dataset.

\cite{zheng2022gatcadnetgraphattentionnetwork} have developed a symbol identification method called "GAT-CADNet: Graph Attention Network for Panoptic Symbol Spotting in CAD Drawings," which approaches the task as a subgraph search problem and leverages attention layers.  They reformulate the symbol spotting task as a subgraph detection problem, solved by predicting the adjacency matrix. They use a relative spatial encoding module to capture relative positional and geometric relationships between vertices. Finally, a cascaded edge encoding module is employed to extract vertex attentions and predict the adjacency matrix.

\cite{9879621} proposes CADTransformer to solve the panoptic segmentation on the FloorplanCAD dataset.  The authors tokenize the graphical primitives in CAD drawings into a set of discrete tokens. To optimize fine-grained semantic and instance predictions, they design a semantic head to predict the categories of graphical primitives and an offset head to shift to their respective ground-truth instance centroids. They incorporate neighborhood-aware self-attention, hierarchical feature aggregation, and graphic entity position encoding. Additionally, they sub-sample layers to create additional training data. 

More recently, \cite{liu2024symbolpointspanopticsymbol} achieved the best results on the FloorplanCAD dataset proposing SymPoint. They propose a novel approach that views graphic primitives as locally connected 2D points and applies point cloud segmentation methods. Their method uses a point transformer \cite{zhao2021pointtransformer} and mask2former spotting head \cite{cheng2022maskedattentionmasktransformeruniversal} and introduces the attention with connection module and contrastive connection learning scheme. They also propose a KNN interpolation mechanism for the mask attention module for the symbol spotting task.

\section{Preliminaries} 
\label{sec:background}
\subsection{Scalable Vector Graphics}
\label{subsec:svg}
\begin{table}
\caption{Path Commands in SVG used in the research}
\label{table:svg-path-commands}
\centering
\small
\renewcommand{\arraystretch}{1.25}
\begin{tabular}{l p{7cm} l}
\hline\hline
\multicolumn{1}{c}{Command} &
\multicolumn{1}{c}{Meaning} &
\multicolumn{1}{c}{Number of Parameters} \\
\multicolumn{1}{c}{(1)} &
\multicolumn{1}{c}{(2)} &
\multicolumn{1}{c}{(3)} \\
\hline
M (moveto) & Move the current point to a new location. & 2 (x, y) \\
L (lineto) & Draw a straight line from the current point to a new location. & 2 (x, y) \\
H (horizontal lineto) & Draw a horizontal line from the current point to a new x-coordinate. & 1 (x) \\
V (vertical lineto) & Draw a vertical line from the current point to a new y-coordinate. & 1 (y) \\
C (curveto) & Draw a cubic Bézier curve from the current point to a new location. & 6 (x1, y1, x2, y2, x, y) \\
S (smooth curveto) & Draw a cubic Bézier curve from the current point to a new location, using the previous control point as the first control point. & 4 (x2, y2, x, y) \\
Q (quadratic curveto) & Draw a quadratic Bézier curve from the current point to a new location. & 4 (x1, y1, x, y) \\
T (smooth quadratic curveto) & Draw a quadratic Bézier curve from the current point to a new location, using the previous control point as the first control point. & 2 (x, y) \\
A (elliptical arc) & Draw an elliptical arc from the current point to a new location. & 
\begin{tabular}{l}
7 (rx, ry, x-axis-rotation, \\[0.2cm]
large-arc-flag, sweep-flag, x, y)  \\

\end{tabular}
\\
Z (closepath) & Close the current path by connecting to the last moveto command. & 0 \\
\hline\hline
\end{tabular}
\normalsize
\end{table}

\begin{figure}[ht]
\centering
\small
\renewcommand{\arraystretch}{1.25}
\begin{tabularx}{\textwidth}{>{\hsize=0.14\hsize}X >{\hsize=0.4\hsize}X >{\hsize=0.4\hsize}X >{\hsize=0.11\hsize}X}
\hline\hline
\multicolumn{1}{c}{Pic} &
\multicolumn{1}{c}{Basic Shape} &
\multicolumn{1}{c}{Path Conversion} &
\multicolumn{1}{c}{Pic2} \\
\hline
\raisebox{-0.5\height}{\includegraphics[scale=0.78]{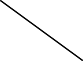}}
& \begin{tabular}{l}
\texttt{<line} \\
\hspace{0.5cm}style="fill:none;stroke:\#000000" \\
\hspace{0.5cm} x1="0" y1="0" \\
\hspace{0.5cm} x2="1" y="1" \\
\texttt{/>}
\end{tabular} & 
\begin{tabular}{l}
\texttt{<path} \\
\hspace{0.5cm}style="fill:none;stroke:\#000000" \\

\hspace{0.5cm} d="m 0,0 L1,1" \\
\texttt{/>}
\end{tabular} & 
\raisebox{-0.5\height}{\includegraphics[scale=0.78]{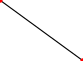}} \\

\raisebox{-0.5\height}{\includegraphics[scale=0.78]{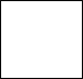}}
& \begin{tabular}{l}
\texttt{<rect} \\
\hspace{0.5cm}style="fill:none;stroke:\#000000" \\
\hspace{0.5cm} width="1" height="1" \\
\hspace{0.5cm} x="0" y="0" \\
\texttt{/>}
\end{tabular} & 
\begin{tabular}{l}
\texttt{<path} \\
\hspace{0.5cm}style="fill:none;stroke:\#000000" \\

\hspace{0.5cm} d="m 0,0 v 1 h 1 v -1 Z" \\
\texttt{/>}
\end{tabular} & 
\raisebox{-0.5\height}{\includegraphics[scale=0.78]{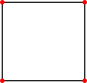}} \\

\raisebox{-0.5\height}{\includegraphics[scale=0.78]{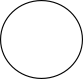}}
& \begin{tabular}{l}
\texttt{<circle} \\
\hspace{0.5cm}style="fill:none;stroke:\#000000" \\
\hspace{0.5cm} r=1 \\
\hspace{0.5cm} cx="0" cy="1" \\
\texttt{/>}
\end{tabular} & 
\begin{tabular}{l}
\texttt{<path} \\
\hspace{0.5cm}style="fill:none;stroke:\#000000" \\

\hspace{0.5cm}d="M 1,0\\
\hspace{0.5cm}A 1,1 0 0 1 0,1\\
\hspace{0.5cm}A 1,1 0 0 1 -1,0\\
\hspace{0.5cm}A 1,1 0 0 1 0,-1\\
\hspace{0.5cm}A 1,1 0 0 1 1,0\\
\hspace{0.5cm}Z" \\
\texttt{/>}
\end{tabular} & 
\raisebox{-0.5\height}{\includegraphics[scale=0.78]{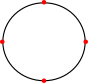}} \\
\hline\hline
\end{tabularx}
\normalsize
\caption{Conversion of basic shapes in paths. Descriptions of path commands are defined in table \ref{table:svg-path-commands}. }
\label{fig:Path_conversion}
\end{figure}

The Portable Document Format (PDF), standardized as ISO 32000 \cite{iso32000}, is a file format for presenting documents in a platform-independent manner. Initially developed by Adobe in 1992, PDFs encapsulate a complete description of a fixed-layout flat document, including text, fonts, vector graphics, and raster images. However, the inherent design of PDFs as a page-description language, focused on describing the layout and appearance of a page rather than the underlying vector data, poses a significant challenge for direct access to vector information.

This limitation results from PDFs being intended as a final-form format, optimized for printing or viewing rather than for editing or manipulating the underlying data. As a result, the vector graphics, including paths, shapes, and text, are often encoded in a manner that complicates the extraction of the original vector data.

The documents in PDF are converted to Scalable Vector Graphics (SVG) to allow vector extraction and manipulation. In contrast to other vector formats, such as Encapsulated PostScript (EPS), the conversion to SVG offers greater accessibility and editability. EPS files are often encoded in a binary code that requires specialized software to read and edit, limiting their flexibility and accessibility. The proprietary nature of EPS also makes it challenging to create software that can accurately parse and manipulate EPS files. In contrast, SVG is an open, text-based standard that can be easily edited using various software tools and programming languages.

The SVG language is an XML-based vector image format for defining two-dimensional graphics, as defined in the SVG 2 specification \cite{w3ScalableVector}. The SVG format is maintained and standardized by the World Wide Web Consortium (W3C), an international community that develops and maintains web standards. 
This format allows for the creation and modification of complex graphical objects, and its open-source representation provides total access to the geometrical data, which is often not possible with vector data formats from proprietary computer-aided drawing tools.

In the context of SVG, a path is the fundamental building block for creating shapes. It consists of a sequence of text-based commands rendered into geometrical shapes according to the format definition. The core concept underlying path commands is the definition of the current point, which is modified to create a shape by tracing a relative geometric line from the current position. Table \ref{table:svg-path-commands} lists the commands used in this study. Although the SVG standard supports more complex commands, such as Circle and Rectangle, which are built upon the path command, we convert them to the basic representation to ensure consistency in representation. Different SVG exports can yield varying representations, such as a rectangle being exported as a rectangle command in one case and as a path of four closing lines in another. 

The conversion of basic shapes to path commands, as illustrated in Figure \ref{fig:Path_conversion}, provides a detailed insight into the underlying structure of SVG graphics. For instance, the representation of a rectangle, depicted in the second row of the table, demonstrates how a simple geometric shape can be decomposed into a sequence of path commands. The original rectangle, defined by the <rect> command with attributes width, height, x, and y, is equivalent to the path command <path> with the d attribute set to "m 0,0 v 1 h 1 v -1 z". This path command sequence can be broken down into individual components: the m 0,0 command moves to the starting point (0,0), the v 1 command draws a vertical line to the point (0,1), the h 1 command draws a horizontal line to the point (1,1), the v -1 command draws a vertical line to the point (1,0), and the Z command closes the path. This decomposition is the core principle underneath the tensor representation, where each decomposed path element is tokenized and used for structured input.


\section{Method} 
\label{sec:method}
\begin{figure}[t]
\centering
\includegraphics[scale=0.365]{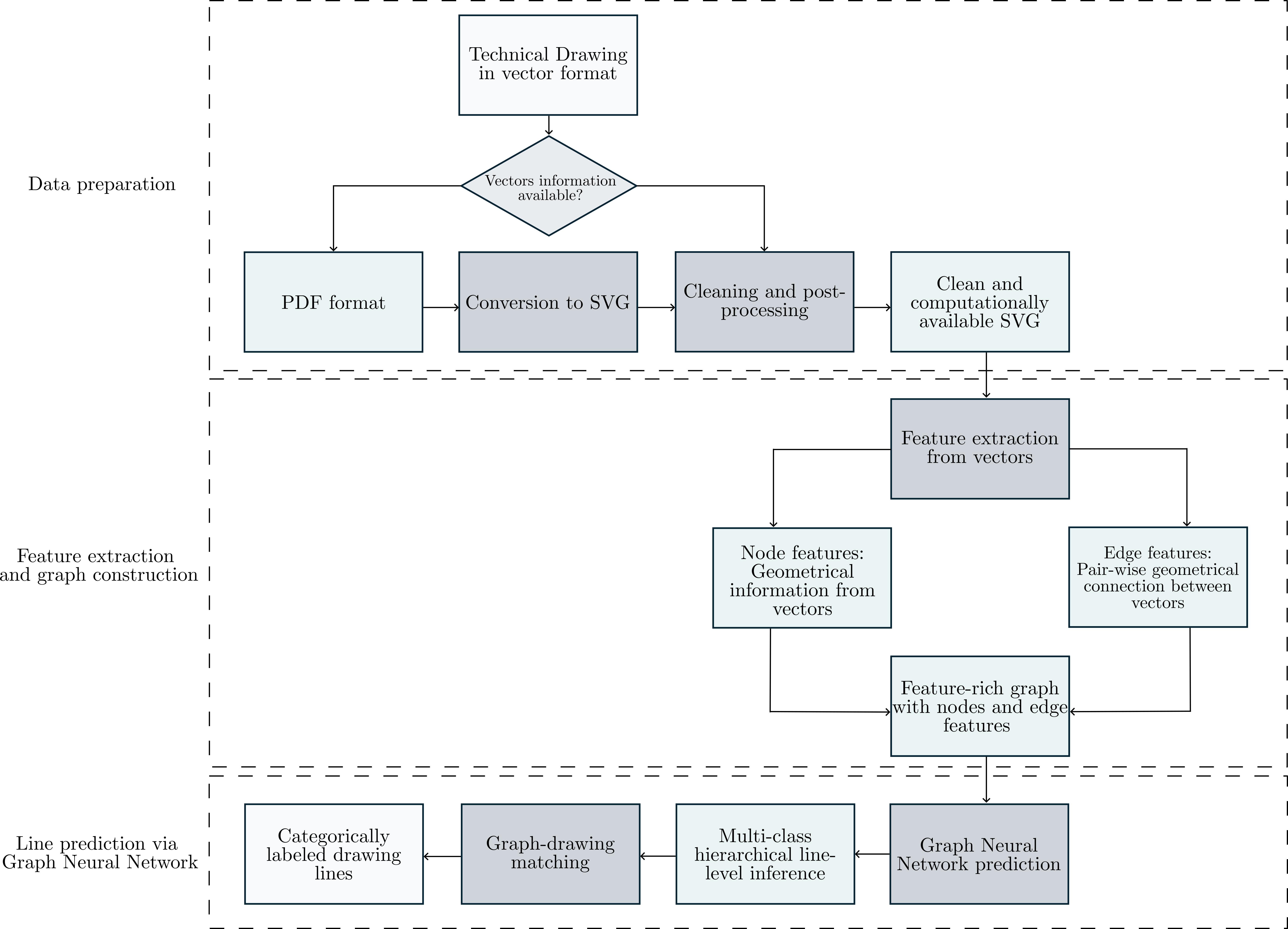}
\caption{Creation process of the graph}
\label{fig:general_flow}
\end{figure}

The objective is to develop an automated method for segmenting line drawings at the line level, utilizing solely the geometric information present in the drawing. The ultimate goal is to achieve a fully automated process that can divide lines into distinct layers, each representing the underlying semantic meaning of the technical drawing.
Our methodology, illustrated in Figure \ref{fig:general_flow}, comprises three primary stages. Initially, the data preparation phase involves two processes: (1) converting PDF files to SVG format and (2) assessing and refining the data to establish a standardized format conducive to analysis. Subsequently, the feature extraction and graph construction stage entails the meticulous extraction of geometrical information from vectorized data, creating a graph that delineates the rich geometrical relationships between vector entities within the drawing. Finally, the third stage involves defining, training, and testing a neural network architecture to perform semantic segmentation.

\subsection{PDF-to-SVG Format Conversion and Data Structuring}
\label{subsec:pdf2svg}
We propose a PDF vector extraction framework that facilitates the systematic extraction and representation of geometric and semantic features, rendering the information computationally accessible. However, the initial SVG output often exhibits a nested, transformation-rich structure, necessitating a dedicated post-processing step to simplify and flatten the geometries, thereby producing a transformation-invariant representation amenable to further analysis.

The process is automated through a combination of vector-specific libraries and rule-based definitions. The first step involves the conversion of PDF to SVG using CairoSVG\footnote{https://cairosvg.org/}, a 2D graphics library that provides a high-level interface for rendering SVG files. We then remove each grouping and transformation at the vector level by defining recursive functions to apply linear transformations and decoupling, thereby eliminating redundant representations.

The outcome of this process is a unique representation of the vector information, wherein each geometric entity is mapped to a distinct and unambiguous representation, eliminating the redundancy introduced by grouping and transformation operations.

\begin{figure}
\centering
\includegraphics[scale=0.75]{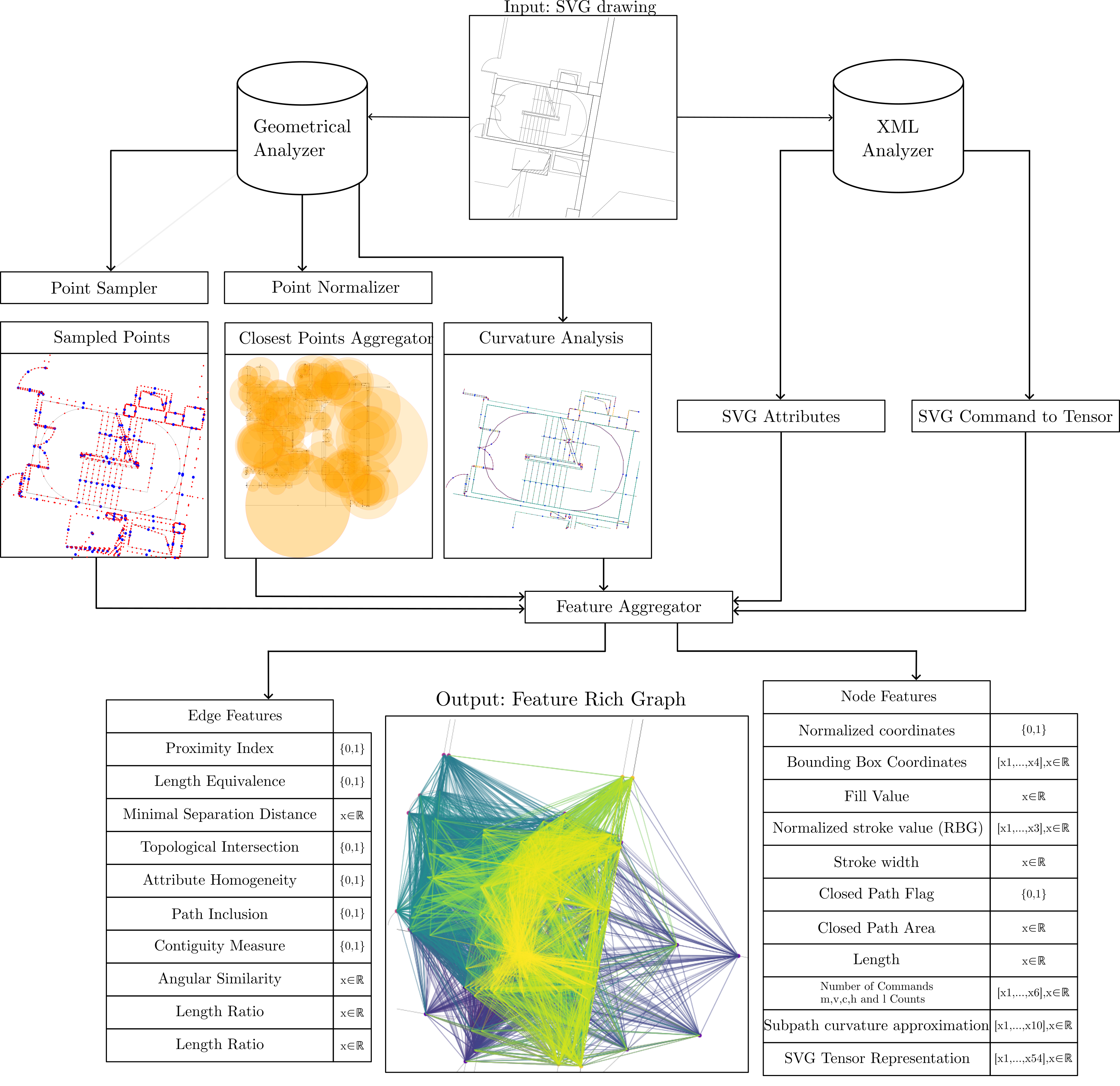}
\caption{Graph creation process: Geometrical and XML analysis are used to extract information, which is then aggregated into edge and node features.}
\label{fig:complex_drawing}
\end{figure}

Following the pre-processing stage, as illustrated in Figure \ref{fig:complex_drawing}, each drawing is transformed into a graph representation. The graph creation process entails two key components: node feature extraction and edge feature extraction. These steps extract relevant information from the drawing, which is then encoded as node and edge attributes in the resulting graph.


\subsection{Node Feature Extraction}
\label{subsec-featext}
In the graph construction process, each path within the SVG is represented as a node, to which we assign both geometrical and stylistic information encoded in the XML definition.

\textbf{Style features.} We parse the XML tags to determine the presence or absence of a fill color, assigning a binary value accordingly. Additionally, we extract the stroke color and stroke width attributes. 

\textbf{Numerical features.} The path is converted into a vectorized format, and its features are extracted through numerical operations. Initially, we determine whether the path is closed, which is represented as a binary value. When a path is closed, its area is calculated. Additionally, we compute the total path length and the total number of commands across all paths. To mitigate the impact of outliers on the training process, we apply a logarithmic transformation to measures susceptible to outliers, such as area and length, thereby reducing their negative influence on the model.
We use a numerical method for estimating the curvature of parametric curves at multiple points along the curve. The approach involves dividing the curve into a series of intervals, with the number of intervals determined by the user-defined parameter $n$. The length of each interval, denoted by $\Delta t$, is calculated as the reciprocal of $n$:

\begin{equation}    
\label{eq:fraction_over}
\Delta t = \frac{1}{n}
\end{equation}

For each interval, we calculate the point on the curve corresponding to the current parameter value $t$, where $t$ ranges from 0 to 1. The point is obtained by evaluating the parametric curve function $p(t)$ at $t$.

To estimate the curvature at each point, we employ a finite difference method, which approximates the second derivative of the curve. Specifically, we calculate the curvature using the formula:

\begin{equation}
\label{eq:curvature_function}
\kappa = \frac{2 \left( (x_0 - x_1) (y_2 - y_1) - (y_0 - y_1) (x_2 - x_1) \right)}{\sqrt{\left( (x_0 - x_1)^2 + (y_0 - y_1)^2 \right)^3}}
\end{equation}
where $(x_0, y_0)$, $(x_1, y_1)$, and $(x_2, y_2)$ are points on the curve, with $(x_1, y_1)$ being the current point, and $(x_0, y_0)$ and $(x_2, y_2)$ being the points at $t - \Delta t$ and $t + \Delta t$, respectively.

We extract the median point to capture the vector's central tendency. This is achieved by sampling $n$ equally spaced points along the path, denoted as $p_1, p_2, ..., p_{n}$
where each point $p_i$ is characterized by its x-coordinate $x_i$ and a y-coordinate $y_i$.

The points are spaced evenly along the path, with a distance of $\Delta s = L/(n-1)$ between each consecutive point, where $L$ is the total length of the path.

We then calculate the median point of these sampled points, which we denote as $\bar{p} = (\bar{x}, \bar{y})$. This median point is taken as the path's representative $(x, y)$ point.

\textbf{SVG Embedding} The final part of the node features is the tensor representation of the SVG path command following the approach presented in DeepSVG. \cite{carlier2020deepsvghierarchicalgenerativenetwork} 

The input data is a string representation of an SVG path, which is a compact notation for describing a sequence of 2D geometric shapes, described in subsection \nameref{subsec:svg}. The string is subjected to lexical analysis of the text describing the geometry in SVG format and tokenization, resulting in a sequence of tokens, denoted as $\mathcal{T} = {t_1, t_2, ..., t_n}$, where each token $t_i$ represents a command or coordinate. The token sequence is then converted into an intermediate representation, a hierarchical data structure that captures the essential information of the SVG path. This intermediate representation can be formalized as a graph $\mathcal{G} = (\mathcal{V}, \mathcal{E})$, where $\mathcal{V}$ is the set of vertices representing commands and coordinates, and $\mathcal{E}$ is the set of edges representing the relationships between them.

\textbf{Tensor representation}

The intermediate representation $\mathcal{I}$ is a class or data structure instance, denoted as $\mathcal{S}$, representing a group of SVG paths. Each path in the group is composed of multiple segments, where each segment is a sequence of points.

Each segment in the path is represented by a tuple of points, denoted as $\mathcal{P} = {(x_1, y_1), (x_2, y_2), ..., (x_m, y_m)}$, where each point $(x_i, y_i)$ is a 2D coordinate.

The intermediate representation $\mathcal{I}$ is then converted into a tensor representation, denoted as $\mathcal{T}$, which is a numerical representation of the path data. The tensor $\mathcal{T}$ is a multi-dimensional array of numbers, where each dimension corresponds to a specific aspect of the path data, such as the command type, coordinates, or other parameters.

The tensor structure can be formalized as a matrix $\mathcal{T} \in \mathbb{R}^{n \times m}$, where $n$ is the number of segments, and $m$ is the number of attributes per segment. Each row of the matrix represents a segment, and each column represents a specific segment attribute.

The SVG path command can comprise an arbitrary number of sub-commands, enabling the creation of complex shapes with varying degrees of intricacy. To ensure consistency in the graph representation, we pad and truncate the resulting tensor to a fixed length of $n$. Tensors with lengths less than $n$ are padded with vectors following the shape of the tensor filled of -1, while those exceeding $n$ are truncated, potentially resulting in information loss. The choice of $n$ is critical, as selecting a value that is too small may lead to excessive information loss, whereas choosing a value that is too large may introduce redundant information for most nodes, potentially impeding or compromising the training of the network.

\subsection{Edge Feature Extraction}
The graph nodes encapsulate the geometric attributes of each line segment. In contrast, the edges embody the topological and semantic correlations between these line segments, capturing the spatial and relational dependencies among them.

\begin{figure}
\centering
\includegraphics[scale=0.4]{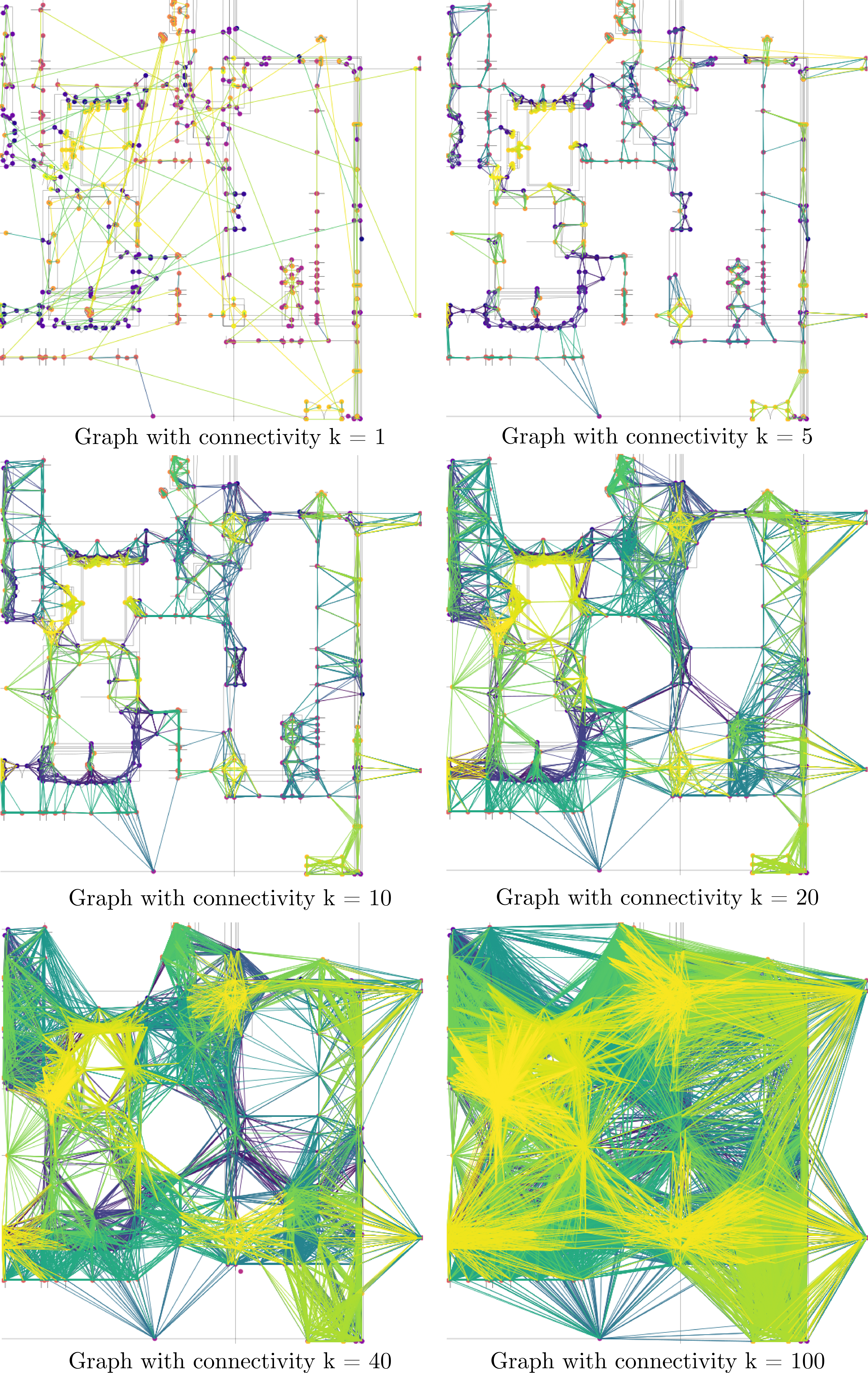}
\caption{Different connectivities on graphs}
\label{fig:graphs_drawings}
\end{figure}

To mitigate the complexity of a fully connected graph, we utilize the K-nearest neighbor (KNN) algorithm to reduce the degree of each node. Specifically, we calculate each node's top K nearest neighbors based on the average vector point obtained from the node-feature extraction process. As illustrated in Figure \ref{fig:graphs_drawings}, the choice of the hyperparameter K significantly impacts the graph's structure: a low value of K results in a sparse graph, while increasing K yields a more densely connected graph.  To ensure the resulting graph is undirected, we remove duplicate appearances of node pairs by aggregating bidirectional edges into a single undirected edge. For instance, any pair of nodes that are connected in both directions are merged into a single undirected connection, effectively halving the number of edges in the graph. This simplification preserves the symmetry of the relationships between nodes while reducing the overall complexity of the graph. This approach is enhanced by adding the possibility of connections between random graph nodes, which encourages the model to learn a more comprehensive representation of the symbols and semantics rather than simply relying on the local context of surrounding nodes, thereby preventing overfitting and poor generalization to new, unseen data, as the model is able to capture long-range dependencies and relationships between nodes that may not be immediately adjacent.
The KNN approach is favored over a distance-threshold cut-off method to guarantee that every node in the graph is connected to at least K other nodes, thereby ensuring a minimum level of connectivity and preventing isolated nodes.

\begin{figure}
\centering
\includegraphics[scale=2]{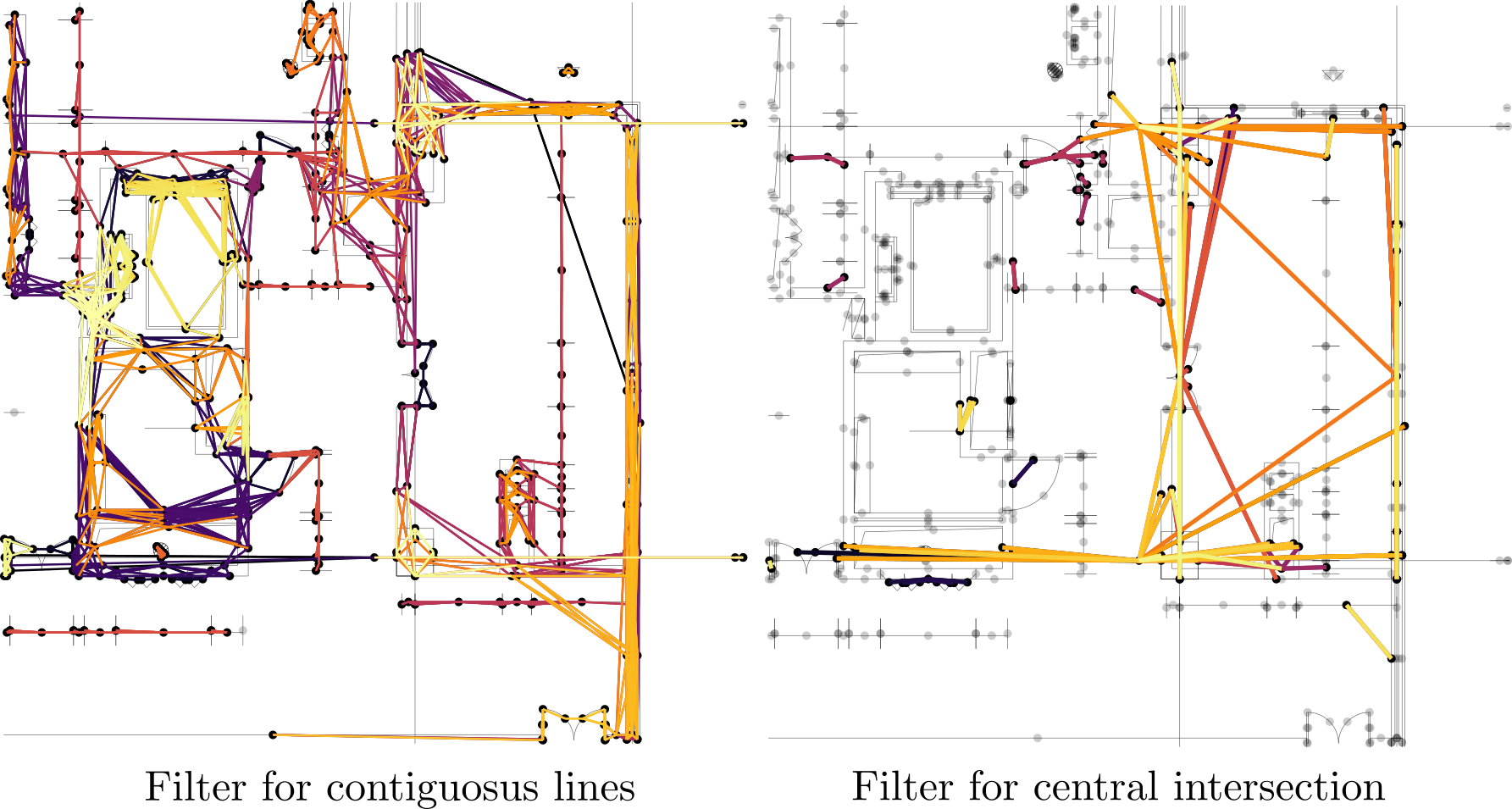}
\caption{Example filtering}
\label{fig:filters}
\end{figure}

The edge feature is a 10-dimensional vector comprising several components that capture the geometric and topological relationships between two segments. The feature information can be used to derive insightful information and filter the types of connections in the graph, for example, Figure \ref{fig:filters} shows the extraction of the contiguous lines and central intersection information from the graph allowing for straightforward application of geometric filters to the drawing.

Specifically, the vector includes two Boolean values indicating whether the segments have identical lengths and whether they are selected based on the vicinity rule or random sampling.
The vector also stores the two segments' logarithmic and inverse length ratios. The normalized angle between the center points of the mean points of the segments, denoted as $p1 = (x1,y1)$ and $p2 = (x2,y2)$, is calculated using the following formula:

\begin{equation}
\label{eq:angle }
\theta_\text{norm} = \frac{\left(\tan^{-1}\left(\frac{\Delta y}{\Delta x}\right) \mod 360\right)}{360}
\end{equation}
where $\Delta x = x_1 - x_2$ and $\Delta y = y_1 - y_2$.

To account for the potential complexity of the vector, we also incorporate the minimum point between the two geometries. This is achieved by representing each geometry using equal-distance sampling and computing the logarithm of the minimum pairwise distance matrix between the points.

In cases where one of the two paths is a closed path, we return a binary feature of whether one path is contained within the other. Furthermore, we include the number of intersections between the two segments, which is approximated using a numerical method on the set of points sampled from the geometry, as described in the subsequent section. This approximation is necessary due to the computational expense of exact intersection calculations for numerous SVG paths.

From the line segments $\mathbf{S}_1$ and $\mathbf{S}_2$ in $\mathbb{R}^2$, parameterized by $t \in [0, 1]$ as follows:

\begin{equation}
\label{eq:parametrised1 }
\mathbf{S}_1(t) = (x_1(t), y_1(t)) = (1-t)\mathbf{P}_1 + t\mathbf{Q}_1
\end{equation}
\begin{equation}
\label{eq:parametrised2 }
\mathbf{S}_2(t) = (x_2(t), y_2(t)) = (1-t)\mathbf{P}_2 + t\mathbf{Q}_2
\end{equation}
where $\mathbf{P}_1, \mathbf{Q}_1, \mathbf{P}_2, \mathbf{Q}_2 \in \mathbb{R}^2$ are the endpoints of the segments.

To detect intersections between $\mathbf{S}_1$ and $\mathbf{S}_2$, we sample the segments at $N$ points each, yielding two sets of points:
\begin{equation}
\label{eq:set1 }
\mathbf{J} = \{\mathbf{S}_1(t_i)\}_{i=0}^{N-1}
\end{equation}
\begin{equation}
\label{eq:set2 }
\mathbf{K} = \{\mathbf{S}_2(t_i)\}_{i=0}^{N-1}
\end{equation}

We then compute the following matrices:

\begin{equation}
\label{eq:matrix1 }
\mathbf{A}_x = \begin{bmatrix}
x_1(t_0) & x_2(t_0) \\
x_1(t_1) & x_2(t_1) \\
\vdots & \vdots \\
x_1(t_{N-1}) & x_2(t_{N-1})
\end{bmatrix}
\end{equation}
\begin{equation}
\label{eq:matrix2 }
\mathbf{A}_y = \begin{bmatrix}
y_1(t_0) & y_2(t_0) \\
y_1(t_1) & y_2(t_1) \\
\vdots & \vdots \\
y_1(t_{N-1}) & y_2(t_{N-1})
\end{bmatrix}
\end{equation}

The intersection points are then detected by solving the following system of linear equations:

\begin{equation}
\label{eq:multiplication_matrix }
\begin{bmatrix}
\mathbf{A}_x & -\mathbf{A}_y
\end{bmatrix} \begin{bmatrix}
t_1 \\
t_2
\end{bmatrix} = \begin{bmatrix}
x_1(t_0) - x_2(t_0) \\
y_1(t_0) - y_2(t_0) \\
\vdots \\
x_1(t_{N-1}) - x_2(t_{N-1}) \\
y_1(t_{N-1}) - y_2(t_{N-1})
\end{bmatrix}
\end{equation}
where $t_1, t_2 \in [0, 1]$ are the parameters of the intersection point.

The solution to this system yields the intersection points, which are then filtered using the following conditions:
\newcommand{\sameSign}{\mathrel{\operatorname{\triangleq}}}

\begin{equation}
\label{eq:Final layers }
\begin{aligned}
\det(\mathbf{A}_x, \mathbf{A}_y) &\neq 0 & \text{non-parallel curves}\\
\det(\mathbf{A}_x, \mathbf{A}_y) &\sameSign (x_1(t_0) - x_2(t_0)) & \text{same sign on x-axis}\\
\det(\mathbf{A}_x, \mathbf{A}_y) &\sameSign (y_1(t_0) - y_2(t_0)) & \text{same sign on y-axis}\\
|\det(\mathbf{A}_x, \mathbf{A}_y)| &\geq |x_1(t_0) - x_2(t_0)| & \text{x-axis distance}\\
|\det(\mathbf{A}_x, \mathbf{A}_y)| &\geq |y_1(t_0) - y_2(t_0)| & \text{y-axis distance}
\end{aligned}
\end{equation}

where $a \sameSign b$ if and only if $a$ and $b$ have the same sign.

A binary flag is incorporated into the feature set when two segments exhibit identical SVG style descriptors, thereby facilitating the identification of stylistically similar elements. Furthermore, the contiguity of the segments is evaluated by comparing their extreme points with a tolerance threshold to account for minor discrepancies arising from potential errors in converting SVG geometries from disparate formats.

\begin{figure}[!ht]
\centering
\includegraphics[scale=10]{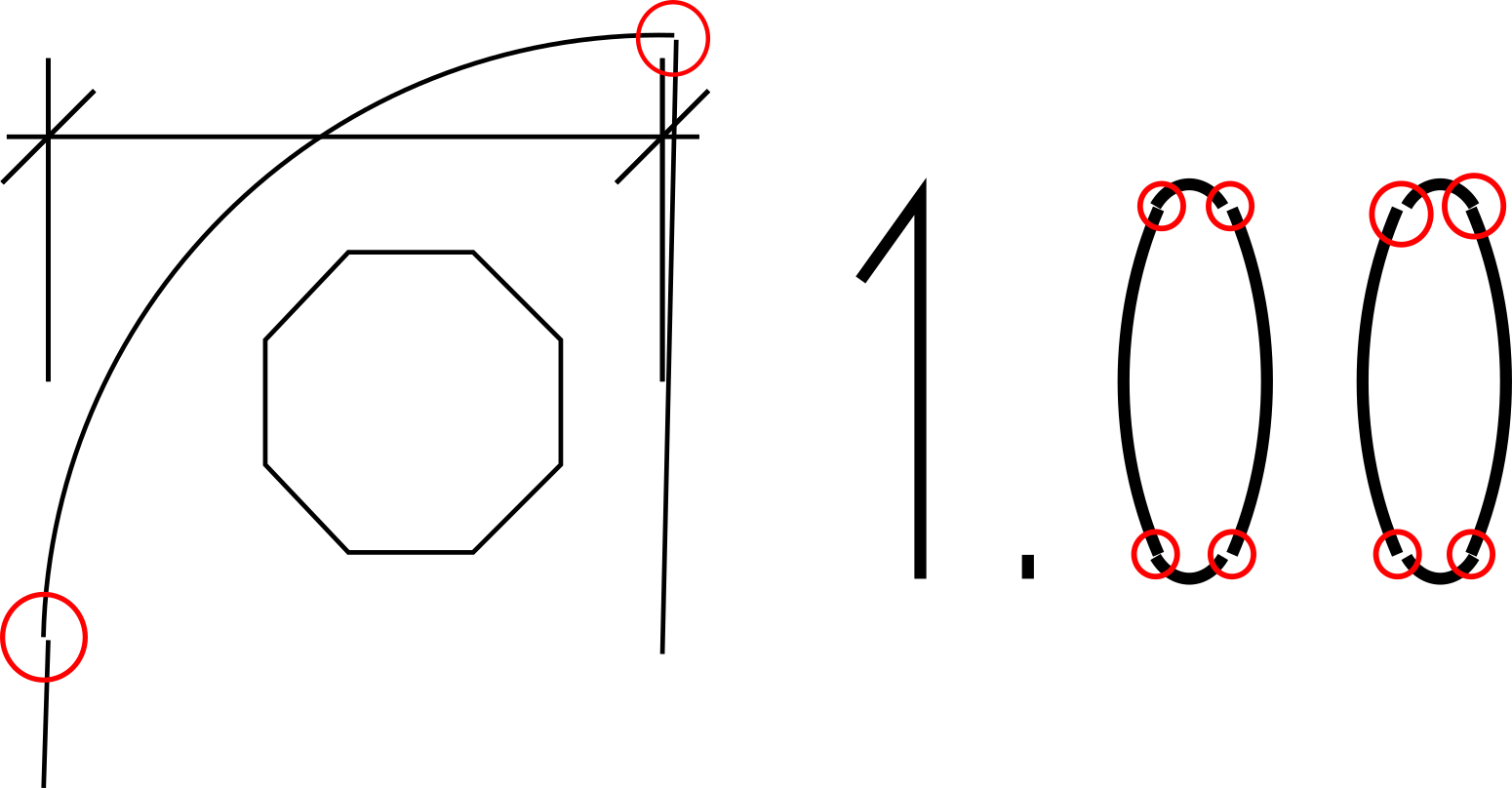}
\caption{Errors in representation}
\label{fig:error_conversion}
\end{figure}

Figure \ref{fig:error_conversion} illustrates examples of such uncertainties, which predominantly occur in text that is not encoded as text in the document but rather converted into geometrical vectors. This discrepancy is noteworthy, as it may not be apparent when examining the raster version of the file, yet it significantly impacts the graph representation. Consequently, correcting these errors is essential to ensure the accuracy and reliability of the graph representation.

\newpage

\subsection{Graph Neural Network}
\begin{figure}
\centering
\scalebox{1}{\includegraphics{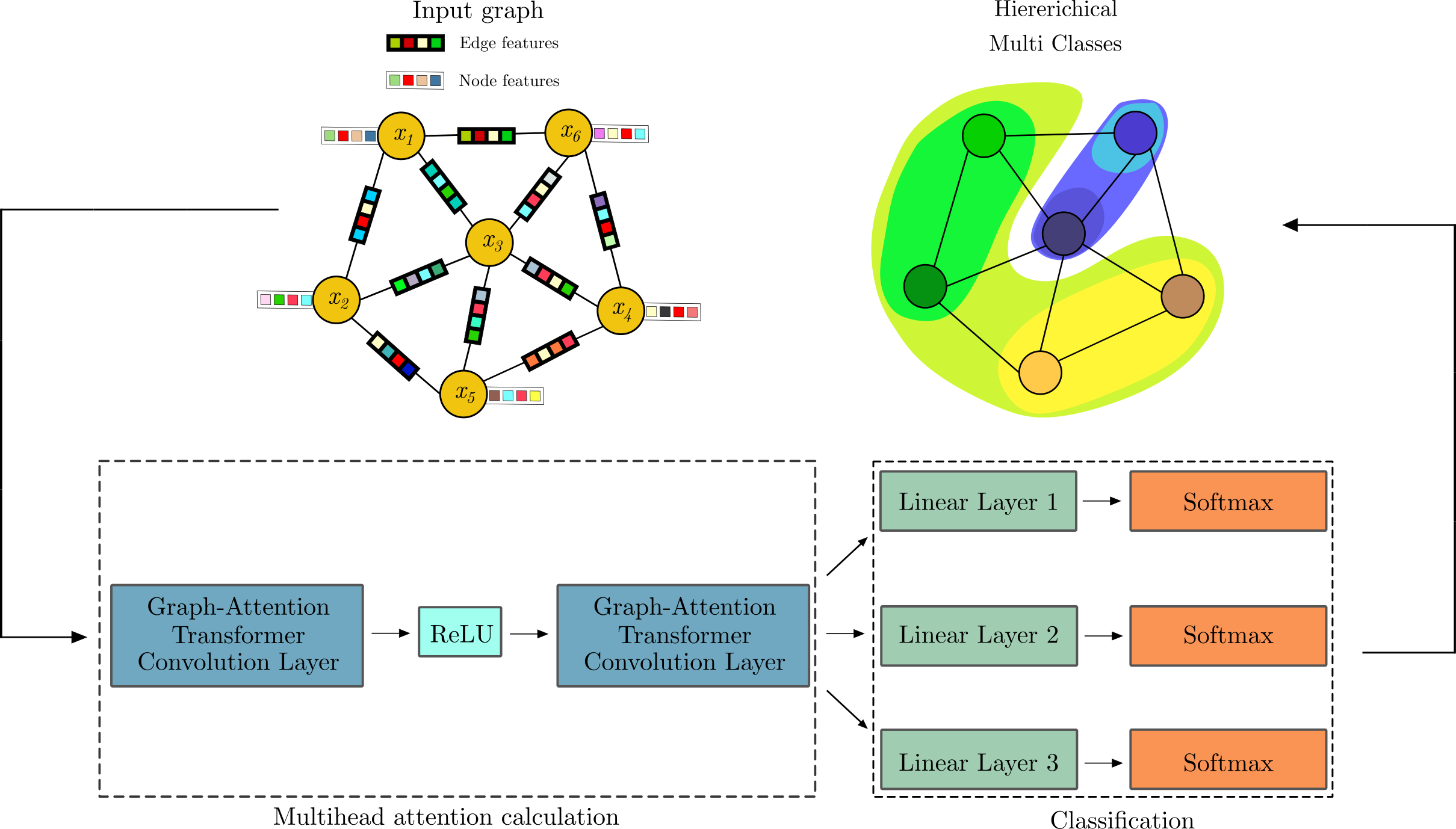}}
\caption{Schematic of the graph transformer architecture }
\label{fig:neural_network}
\end{figure}

To effectively benefit from the complex relationships between nodes in the graph, we propose a neural network architecture incorporating attention mechanisms to selectively focus on the most relevant neighbors, illustrated in Figure \ref{fig:neural_network}. This approach enables the network to learn a weighted representation of the graph structure, where the importance of each neighbor is adaptively determined. Our model architecture is based on this principle and consists of a sequence of two Graph Attention v2 layers \cite{brody2021attentive} interspersed with a ReLU activation layer. The Graph Attention Layer updates each node's features, gathering information from its neighbors. This is weighted by adaptive learned coefficients, allowing the network to learn to select the most important neighbors following the formula:
\begin{equation}
\label{eq:example}
\mathbf{x}_i^{\prime} = \alpha_{i, i} \boldsymbol{\Phi}_s \mathbf{x}_i + \sum_{j \in \mathcal{N}(i)} \alpha_{i, j} \boldsymbol{\Phi}_t \mathbf{x}_j
\end{equation}
The node's feature vector is defined by $\mathbf{x}_i^\prime$, and this refinement is executed through data aggregation from adjacent nodes, employing aggregation weights defined by attention coefficients.

Considering a node $i$ with its feature vector $\mathbf{x}_i^\prime$. The node's intrinsic features transform a learnable weight matrix $\mathbf{\Phi}_s$, and are then scaled by its self-attention coefficient $\alpha_{i,i}$. Parallelly, the feature vectors $\mathbf{x}_j$ of each adjacent node $j$ are altered through a distinct learnable weight matrix $\mathbf{\Phi}_t$, and subsequently scaled by the attention coefficient $\alpha_{i,j}$. The new feature vector $\mathbf{x}_i^\prime$ results from the joint sum of these modified features, including the node and all its neighbors.

\begin{equation}
\label{eq:example1}
\alpha_{i, j}=\frac{\exp \left(\mathbf{a}^{\top} \operatorname{LeakyReLU}\left(\boldsymbol{\Phi}_s \mathbf{x}_i+\boldsymbol{\Phi}_t \mathbf{x}_j+\boldsymbol{\Phi}_e \mathbf{e}_{i, j}\right)\right)}{\left.\sum_{k \in \mathcal{N}(i) \cup\{i\}} \exp \left(\mathbf{a}^{\top} \operatorname{LeakyReLU}\left(\boldsymbol{\Phi}_s \mathbf{x}_i+\boldsymbol{\Phi}_t \mathbf{x}_k+\boldsymbol{\Phi}_e \mathbf{e}_{i, k}\right]\right)\right)}
\end{equation}

The degree of influence that every neighbor, including the node itself, has on the updated feature vector is determined mostly by the attention coefficients.

Using the properties of both nodes, the attention coefficient $\alpha_{i,j}$ is inferred for a pair of nodes, $i$ and $j$. The weight matrices $\mathbf{\Phi}_s$ and $\mathbf{\Phi}_t$ of node $i$ and its neighbor $j$ initially alter its attributes. These modified features are combined and passed through a LeakyReLU activation function to add non-linearity. Furthermore, the edge characteristics between nodes $i$ and $j$ are considered by integrating an edge-specific term $\mathbf{\Phi}_e\mathbf{e}_{i,k}$. This combination is translated into a scalar value by a trainable vector $\mathbf{a}$, which is then exponential.

Normalizing these exponential values across all the node $i$ neighbors (including $i$) generates the final attention coefficient, ensuring a cumulative sum of 1.

The output of the second Graph Attention v2 layer is fed into a multi-class hierarchical classification module, which is designed to predict the class labels for each line in the drawing. This module consists of three parallel linear layers, each followed by a softmax activation function, to produce probabilistic outputs for each individual class. The classes in this context represent the hierarchical categories to which the lines can belong. These are derived from the layer summary of the CAD drawing and annotated with semantic labels. These labels denote the functional role of each element within the drawing. The hierarchy definition is explained in the subsection \nameref{subsec:label}.

The primary classification task involves predicting whether a given line corresponds to one of the over-represented labels in the dataset or a less common category. The second stage involves intermediate categorization, where lines are distinguished between general categories and more specific representations. The third and final stage represents each line's precise and detailed classification, providing a fine-grained categorization that distinguishes between subtly different features within the specific categories.

The weights for each classification task are shared across the hierarchical structure, enabling the model to leverage common patterns and features that are relevant across different classes. This shared-weight approach allows the model to capture class relationships and improve its overall performance. By doing so, the model can effectively identify the most relevant class labels for each line, which is essential for achieving accurate node classification and graph understanding. The hierarchical classification approach also helps to mitigate the issue of class imbalance, where certain classes may be underrepresented in specific subsets of the dataset."

\section{Experimental Setup} \label{sec:experimental_setup}
We present the experimental setup of our approach, which was evaluated using two distinct datasets: a proprietary dataset comprising technical drawings from the Technical University of Munich and the publicly available FloorPlanCAD dataset.\cite{fan2021floorplancadlargescalecaddrawing}.

The proprietary dataset from the Technical University of Munich (TUM) consists of a comprehensive collection of floor plans for 10 different buildings on the university's main campus, each with multiple floors and diverging construction years. The dataset includes a total of 76 floor plans, representing a diverse range of floor layouts and building structures. The floor plans are in SVG format and contain detailed information about the location of rooms, corridors, staircases, and other features.

The dataset covers various building types and floor layouts, including basements, ground floors, and upper floors. Each building has a unique set of floor plans, with some buildings having as few as 3-4 floors and others having up to 6-7 floors. The dataset includes floor plans for buildings with different architectural styles and layouts, such as buildings with a central staircase, buildings with multiple wings, and buildings with a mix of open spaces and private offices.

\subsection{TUM Dataset}
\begin{figure}
\centering
\includegraphics[scale=0.6]{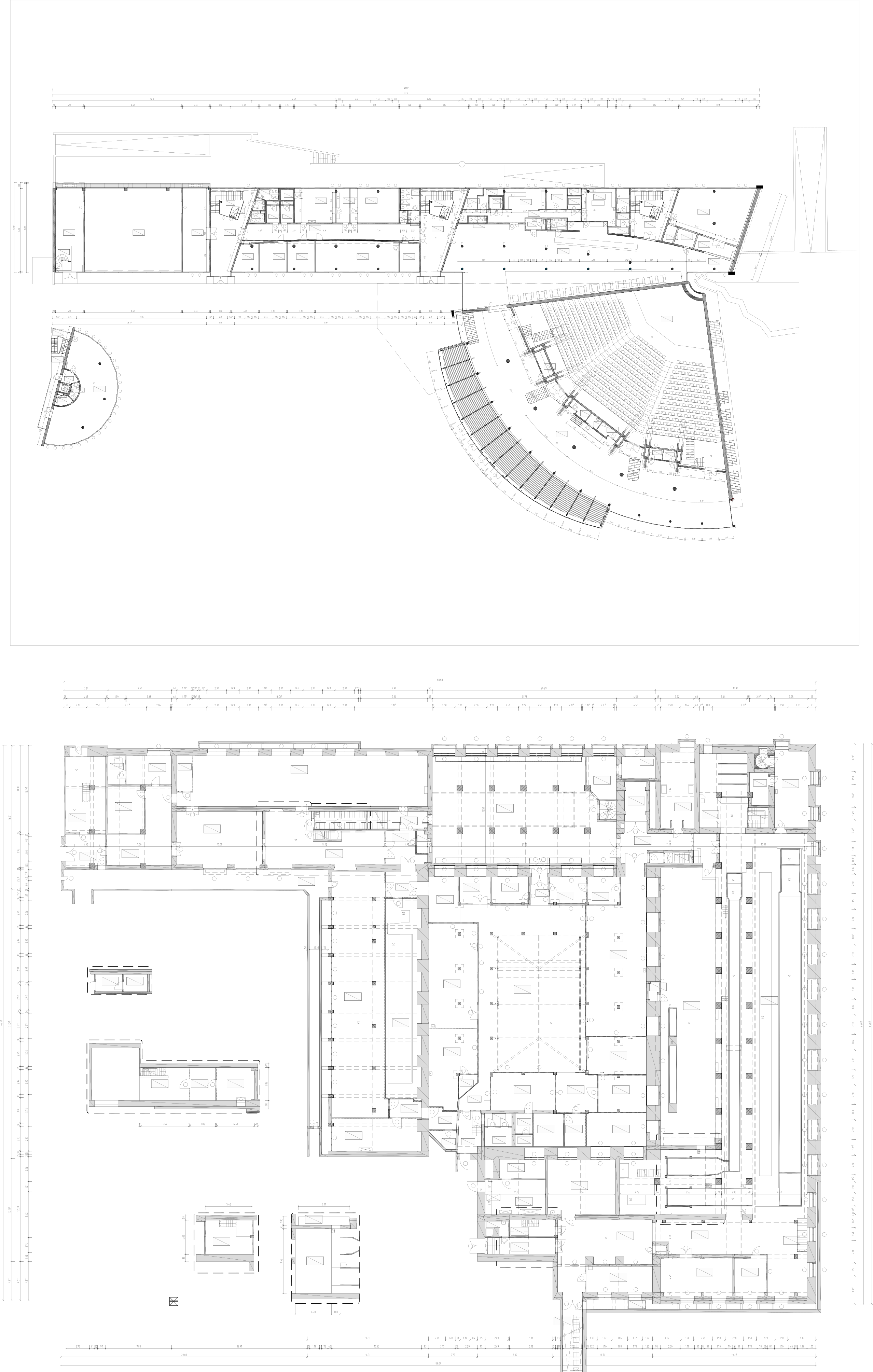}
\caption{Example of drawings in the TUM dataset}
\label{fig:Examples_dataset}
\end{figure}

The internal dataset is obtained from CAD files that are exported in PDF format and undergo our conversion process to SVG format. This process ensures that the method works for each possible geometric file and does not require only the vector format directly from CAD drawing software. In the pre-processing phase, we filter out the text from the technical drawings to ensure the learning from only geometrical information. 
The TUM Dataset consists of 76 Scalable Vector Graphics (SVG) files, with sample images shown in Figure \ref{fig:Examples_dataset}. The minimum number of lines observed in a single drawing is 1,000, while the mean number of lines is approximately 13,500. Furthermore, the maximum number of lines present in a single drawing reaches 34,000.
The architectural drawings comprise a dataset of building information from various structures on the main campus of the Technical University of Munich. The drawings are organized according to the floor plan of each building, providing a representation of the architectural features. Specifically, the drawings include structural elements, such as walls, columns, and beams, and hatching patterns to indicate wall textures and materials. Additionally, the drawings feature representations of architectural elements, including door and window placements, room layouts, and dimensional measurements of the components.

\subsection{Labels extraction and hierarchy definition} \label{subsec:label}
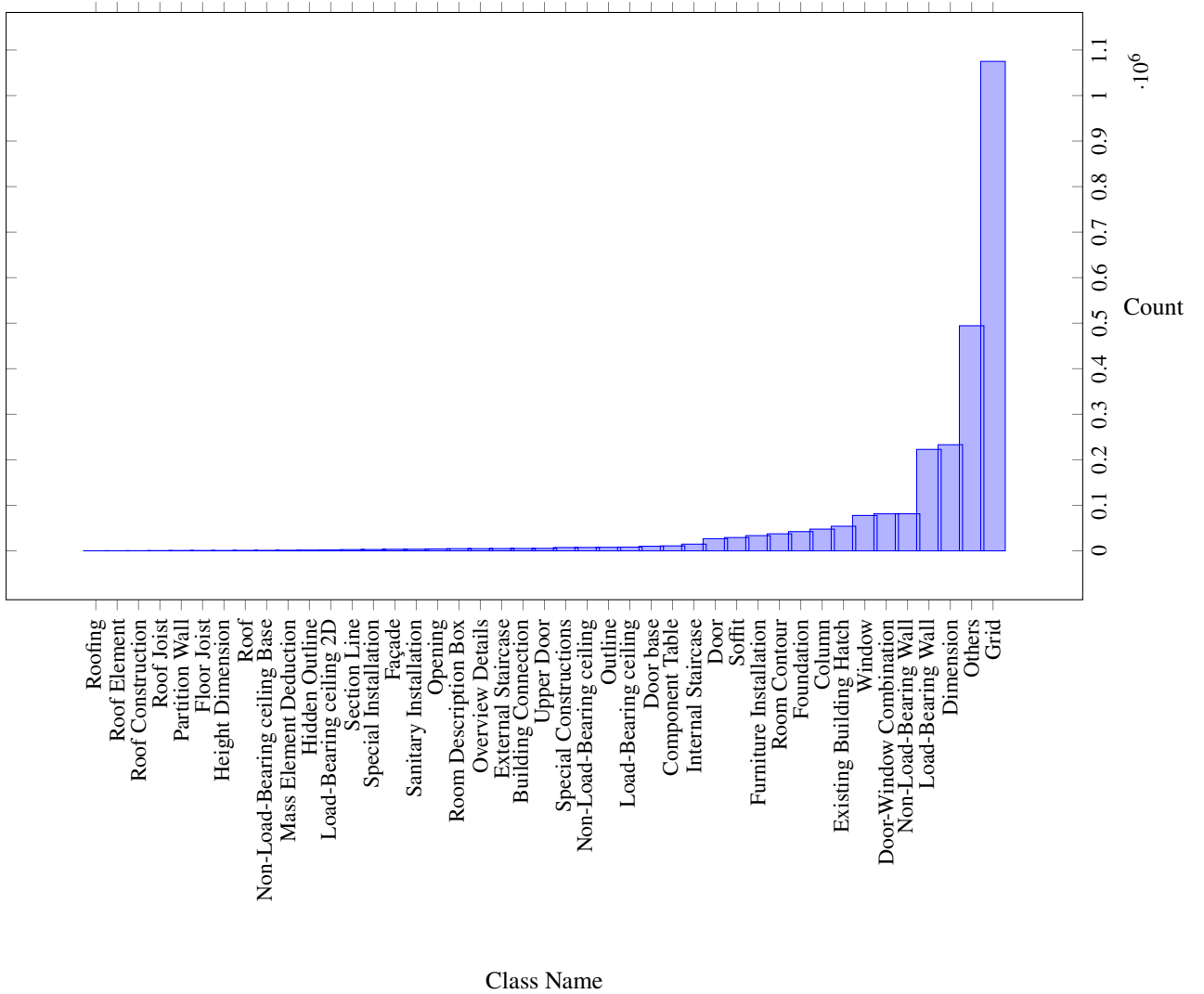
\begin{figure}
\rotatebox{90}{
\begin{tikzpicture}
\begin{axis}[
    xbar,
    width=10cm,
    height=17cm,
    xlabel={\rotatebox{270}{Count}},
    ylabel={\rotatebox{180}{Class Name}},
    ylabel style={yshift=2.2em},
    xticklabel style={font=\footnotesize},
    yticklabel style={font=\footnotesize},
    xtick={0,100000,200000,300000,400000,500000,600000,700000,800000,900000,1000000,1100000,1200000},
    ytick=data,
    yticklabels={
        Grid,
        Others,
        Dimension,
        Load-Bearing Wall,
        Non-Load-Bearing Wall,
        Door-Window Combination,
        Window,
        Existing Building Hatch,
        Column,
        Foundation,
        Room Contour,
        Furniture Installation,
        Soffit,
        Door,
        Internal Staircase,
        Component Table,
        Door base,
        Load-Bearing ceiling,
        Outline,
        Non-Load-Bearing ceiling,
        Special Constructions,
        Upper Door,
        Building Connection,
        External Staircase,
        Overview Details,
        Room Description Box,
        Opening,
        Sanitary Installation,
        Façade,
        Special Installation,
        Section Line,
        Load-Bearing ceiling 2D,
        Hidden Outline,
        Mass Element Deduction,
        Non-Load-Bearing ceiling Base,
        Roof,
        Height Dimension,
        Floor Joist,
        Partition Wall,
        Roof Joist,
        Roof Construction,
        Roof Element,
        Roofing
    }
]
\addplot coordinates {
    (1074982,0)
    (494421,1)
    (233065,2)
    (222782,3)
    (81404,4)
    (81312,5)
    (77727,6)
    (54008,7)
    (47730,8)
    (42298,9)
    (37417,10)
    (33417,11)
    (29209,12)
    (26636,13)
    (14762,14)
    (10883,15)
    (10126,16)
    (8213,17)
    (7848,18)
    (7459,19)
    (7340,20)
    (5421,21)
    (5183,22)
    (5017,23)
    (4835,24)
    (4805,25)
    (4200,26)
    (3537,27)
    (3451,28)
    (2906,29)
    (2736,30)
    (2016,31)
    (1701,32)
    (1198,33)
    (978,34)
    (955,35)
    (714,36)
    (695,37)
    (662,38)
    (474,39)
    (257,40)
    (176,41)
    (119,42)
};
\end{axis}
\end{tikzpicture}}
\caption{Label distribution in TUM dataset}
\label{fig:Label distribution}
\end{figure}

The graphical representation is annotated with semantic labels derived from the layer summary of the CAD drawing. These labels, represented in \ref{fig:Label distribution}, denote the functional role of each element within the drawing. The original names of the labels are in German, represented in column 6 of \ref{table:annotations-classes}; we translate them into English (column 7 of \ref{table:annotations-classes}), and we post-process the list of semantic labels through rule definition.

The layer structure used in the CAD drawing is based on the Bavarian Procurement Manual for Professional Services (VHF Bayern), specifically section VI.4.2.H, which defines the standard for layer organization \cite{bayern}.

To mitigate the issue of class imbalance, where certain classes may be underrepresented in specific subsets of the dataset, we aggregate layer classifications that occur in fewer than 33\% of the total drawings in the class "Others." This strategy enables the model to learn a more generalized representation rather than overfitting to specific drawing types.

The label distribution in the dataset, as illustrated in Figure \ref{fig:Label distribution}, exhibits a pronounced class imbalance, with the "grid" and "others" classes being disproportionately represented. To solve this issue, we employ a multi-class hierarchical learning approach, wherein the primary classification task involves predicting whether a given line corresponds to the "grid," "others," or one of the remaining labels in the dataset.
The second class of labels is defined as intermediate categorization (column 8 in \ref{table:annotations-classes}) of each line in the drawing, wherein we distinguish between two primary categories: annotation lines and effective architectural elements. The annotation lines refer to lines that provide supplementary information or annotations to the drawing, like dimension lines, leader lines, or text annotations. In contrast, architectural elements comprise lines representing actual physical components of the building or structure, including walls, doors, windows, and other architectural features. In the second stage, we group hierarchically  based on semantic similarity, categorizing related elements into higher-level classes. For instance, we aggregate "door," "window," and "opening" into a single higher-level class denoted as "Windows and openings." 

Ultimately, the third and final class represents each element's precise and detailed classification, providing a fine-grained categorization that distinguishes between subtly different architectural features, defined as a numerical class in column 3 and as element description in columns 6 and 7 of table \ref{table:annotations-classes}.\\

\begin{table}[ht]
\caption{Annotations and their corresponding classes in the TUM dataset}
\label{table:annotations-classes}
\centering
\small
\resizebox{\textwidth}{!}{\begin{tabular}{p{0.3cm} p{0.3cm} p{0.3cm} p{1.5cm} p{2cm} p{6cm} p{5cm} p{4cm} l}
\hline\hline
\multicolumn{1}{c}{Class 1} &
\multicolumn{1}{c}{Class 2} &
\multicolumn{1}{c}{Class 3} &
\multicolumn{1}{c}{Percentage} &
\multicolumn{1}{c}{Absolute} &
\multicolumn{1}{c}{Name} &
\multicolumn{1}{c}{Translation} & 
\multicolumn{1}{c}{Intermediate categorization} \\
\multicolumn{1}{c}{(1)} &
\multicolumn{1}{c}{(2)} &
\multicolumn{1}{c}{(3)} &
\multicolumn{1}{c}{(4)} &
\multicolumn{1}{c}{(5)} &
\multicolumn{1}{c}{(6)} &
\multicolumn{1}{c}{(7)} &
\multicolumn{1}{c}{(8)} \\
\hline
2 & 11 & 1 & 0,4049 & 1074982 & A\_00\_RASTER & Grid & Annotations \\
1 & 0 & 13 & 0,0878 & 233065 & A\_03\_BEMASS & Dimension & Annotations \\
1 & 0 & 14 & 0,0003 & 714 & A\_03\_HOEHENKOTE & Height Dimension & Annotations \\
1 & 0 & 18 & 0,0030 & 7848 & A\_08\_UMRISS & Outline & Annotations \\
1 & 0 & 19 & 0,0006 & 1701 & A\_08\_UMRISS\_VERDECKT & Hidden Outline & Annotations \\
1 & 0 & 30 & 0,0203 & 54008 & A\_19\_BESTAND\_SCHR & Existing Building Hatch & Annotations \\
1 & 0 & 31 & 0,0020 & 5183 & A\_19\_GEBAEUDEANSCHLUSS & Building Connection & Annotations \\
1 & 0 & 37 & 0,0141 & 37417 & A\_301\_ADTRAUM & Room Contour & Annotations \\
1 & 0 & 38 & 0,0018 & 4805 & A\_301\_ADTRAUM\_BS & Room Description Box & Annotations \\
1 & 0 & 39 & 0,0041 & 10883 & A\_401\_BAUTEILTABELLE & Component Table & Annotations \\
1 & 0 & 40 & 0,0005 & 1198 & A\_431\_MASSENELEMENT\_ABZUG & Mass Element Deduction & Annotations \\
1 & 0 & 41 & 0,0010 & 2736 & A\_60\_SCHNITTLINIE & Section Line & Annotations \\
1 & 0 & 42 & 0,0018 & 4835 & A\_61\_UEBERSICHT\_DETAILS & Overview Details & Annotations \\
1 & 7 & 21 & 0,0031 & 8213 & A\_09\_TRAGDECKE & Load-Bearing ceiling & Ceiling \\
1 & 7 & 22 & 0,0008 & 2016 & A\_09\_TRAGENDE\_BAUTEILE\_2D & Load-Bearing ceiling 2D & Ceiling \\
1 & 7 & 23 & 0,0110 & 29209 & A\_10\_UNTERZUG & Soffit & Ceiling \\
1 & 7 & 24 & 0,0028 & 7459 & A\_11\_LEICHTDECKE & Non-Load-Bearing ceiling & Ceiling \\
1 & 7 & 25 & 0,0004 & 978 & A\_11\_LEICHTDECKE\_BS & Non-Load-Bearing ceiling Base & Ceiling \\
1 & 3 & 9 & 0,0180 & 47730 & A\_014\_STUETZE & Column & Columns Foundations \\
1 & 3 & 10 & 0,0159 & 42298 & A\_015\_FUNDAMENT & Foundation & Columns Foundations \\
1 & 4 & 15 & 0,0013 & 3451 & A\_05\_FASSADE & Façade & Façade \\
1 & 10 & 34 & 0,0126 & 33417 & A\_26\_MOEBEL\_EINBAU & Furniture Installation & Installations \\
1 & 10 & 35 & 0,0011 & 2906 & A\_27\_EINBAU\_BESOND & Special Installation & Installations \\
1 & 10 & 36 & 0,0013 & 3537 & A\_28\_EINBAU\_SANIT & Sanitary Installation & Installations \\
1 & 9 & 32 & 0,0002 & 474 & A\_22\_AUSSPAR\_DECKE & Roof Joist & Joist \\
1 & 9 & 33 & 0,0003 & 695 & A\_23\_AUSSPAR\_BODEN & Floor Joist & Joist \\
1 & 8 & 26 & 0,0004 & 955 & A\_12\_DACH & Roof & Roof \\
1 & 8 & 27 & 0,0001 & 176 & A\_121\_DACHELEMENT & Roof Element & Roof \\
1 & 8 & 28 & 0,0001 & 257 & A\_122\_DACHKONSTRUKTION & Roof Construction & Roof \\
1 & 8 & 29 & 0,0000 & 119 & A\_123\_DACHBELAG & Roofing & Roof \\
1 & 6 & 20 & 0,0028 & 7340 & A\_09\_3D\_SONDER & Special Constructions & Special \\
1 & 5 & 16 & 0,0019 & 5017 & A\_06\_TREPPE & External Staircase & Stairs \\
1 & 5 & 17 & 0,0056 & 14762 & A\_06\_TREPPE\_2D & Internal Staircase & Stairs \\
0 & 0 & 0 & 0,1862 & 494421 & Others & Others & Undefined \\
1 & 1 & 2 & 0,0839 & 222782 & A\_01\_TRAGWAND & Load-Bearing Wall & Walls \\
1 & 1 & 11 & 0,0307 & 81404 & A\_02\_LEICHTWAND & Non-Load-Bearing Wall & Walls \\
1 & 1 & 12 & 0,0002 & 662 & A\_021\_TRENNWAND & Partition Wall & Walls \\
1 & 2 & 3 & 0,0293 & 77727 & A\_011\_FENSTER & Window & Windows and openings \\
1 & 2 & 4 & 0,0016 & 4200 & A\_012\_OEFFNUNG & Opening & Windows and openings \\
1 & 2 & 5 & 0,0100 & 26636 & A\_013\_TUER & Door & Windows and openings \\
1 & 2 & 6 & 0,0038 & 10126 & A\_013\_TUER\_BS & Door base & Windows and openings \\
1 & 2 & 7 & 0,0306 & 81312 & A\_013\_TUER\_FENSTER\_KOMBI & Door-Window Combination & Windows and openings \\
1 & 2 & 8 & 0,0020 & 5421 & A\_013\_TUER\_OBERHALB & Upper Door & Windows and openings \\
\hline\hline
\end{tabular}}

\end{table}

\newpage
During the vector graphics conversion process, inaccuracies or discrepancies can emerge in the dataset, particularly in the numerical values associated with dimension lines and symbolic representations represented in Figure \ref{fig:error_conversion}. To mitigate this issue, we account for potential gaps in the calculation of connectivity rules governing the graph generation process, thereby ensuring the fidelity of the resulting dataset.

\subsection{FloorplanCAD dataset}
The flooplanCAD dataset \cite{fan2021floorplancadlargescalecaddrawing} constitutes a large-scale, real-world CAD drawing dataset comprising over 10,000 floorplans, encompassing a diverse range of residential and commercial buildings. The primary objective of this dataset is panoptic symbol detection, which involves not only detecting countable objects such as furniture, windows, and doors in CAD drawings but also establishing the semantics of uncountable elements, including walls.

The dataset is presented as patches, which are partial segments of larger technical drawings. Each patch provides a localized view of the original drawing. 

One notable limitation of the floorplanCAD dataset is the absence of labels for all the types of lines in the drawing. To address this limitation, we adopt the hierarchical grouping strategy based on semantic similarity, analogous to the approach employed in the TUM dataset. The first class involves distinguishing between the symbol's lines and others. The second class is again the super-group of elements having similar semantic definitions, and the third class is the precise classification of the symbol.

\section{Results} 
\label{sec:results}
\subsection{Results on FloorplanCAD dataset}
We compare the training outcomes using the hierarchical multi-label definition proposed in the study against a simplified architecture featuring a single-label prediction specific to the class of interest. The experiments are conducted under identical conditions, with the same hyperparameter configuration, batch size, and dataset partitions (training, testing, and validation sets) to ensure a fair comparison.

The primary objective of this investigation is to empirically validate the efficacy of formulating the learning algorithm as a multi-class task, thereby assessing the benefits of incorporating hierarchical multi-label definitions in the classification paradigm.

A comparative analysis of the classification performance reveals that the multi-label hierarchical approach outperforms the simple class prediction approach in both macro and weighted average metrics, as reported in Table \ref{table:combarison_hierarchical}.

Specifically, the multi-label hierarchical approach achieves a macro average precision of 0.82, recall of 0.76, and F1-score of 0.79. In contrast, the simple class prediction approach yields a macro average precision of 0.74, recall of 0.64, and F1-score of 0.67. This represents a relative improvement of 10.8\% in precision, 19\% in recall, and 18\% in F1-score for the multi-label hierarchical approach.

\begin{figure}
\centering
\includegraphics[scale=1.78]{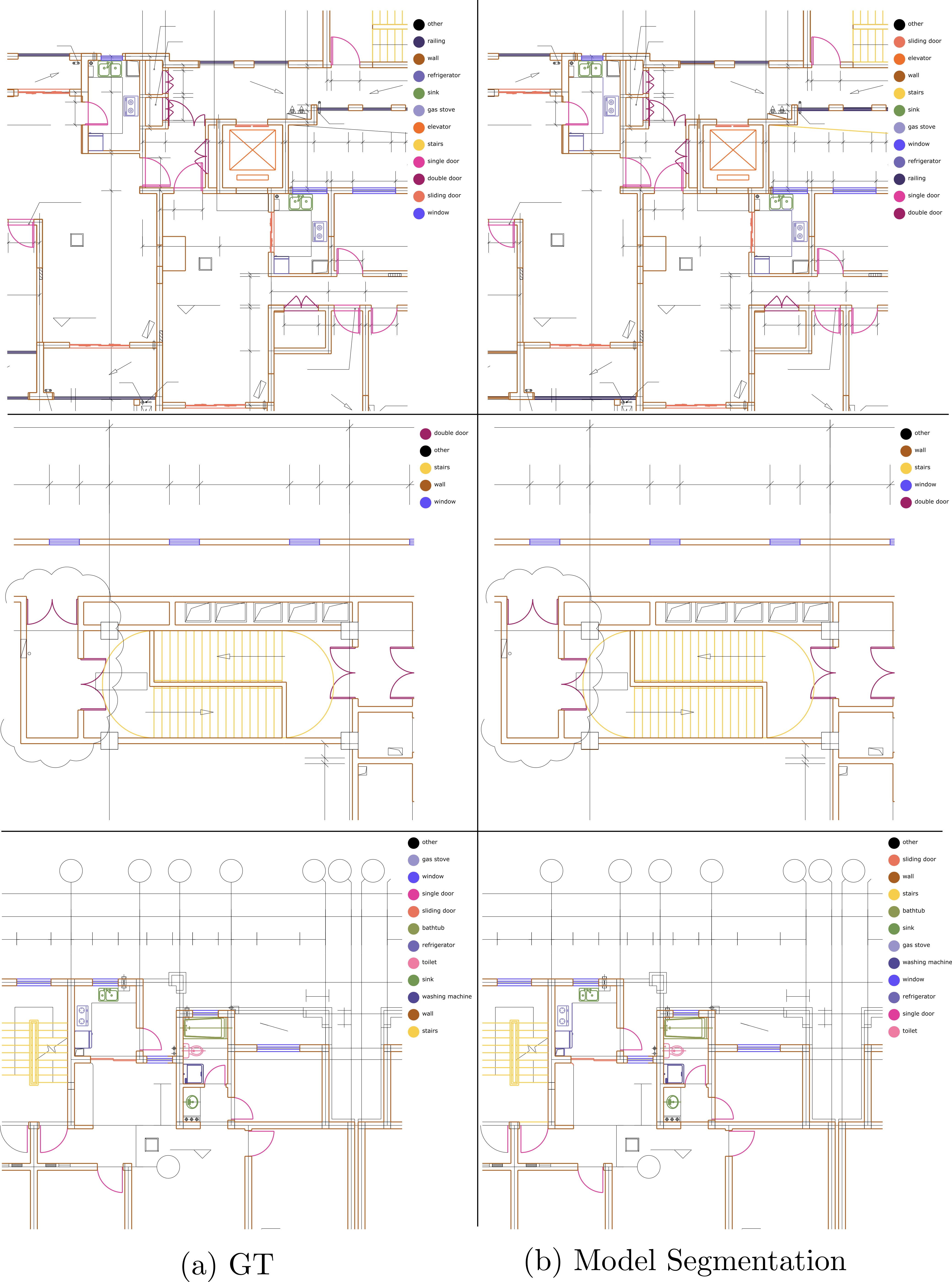}
\caption{Results on FloorplanCAD on line-level segmentation. The left part of the image represents the ground truth, and the right part represents the model segmentation.}
\label{fig:Results_FloorplanCAD}
\end{figure}

Similarly, the weighted average metrics demonstrate superior performance for the multi-label hierarchical approach, with a precision of 0.89, recall of 0.89, and F1-score of 0.88, which is equivalent to the simple class prediction approach, also with a precision of 0.87, recall of 0.87, and F1-score of 0.87.

These findings indicate that the multi-label hierarchical approach is more effective in capturing the underlying patterns in the data, leading to improved classification performance compared to the simple class prediction approach. Notably, the hierarchical approach demonstrates particular strength in learning to represent underrepresented classes in the dataset, suggesting its potential to mitigate class imbalance issues and improve overall model robustness. Visual comparison of VectorGraphNet and ground truth is displayed in Figure \ref{fig:Results_FloorplanCAD}.

\begin{figure}
\centering
\includegraphics[scale=0.74]{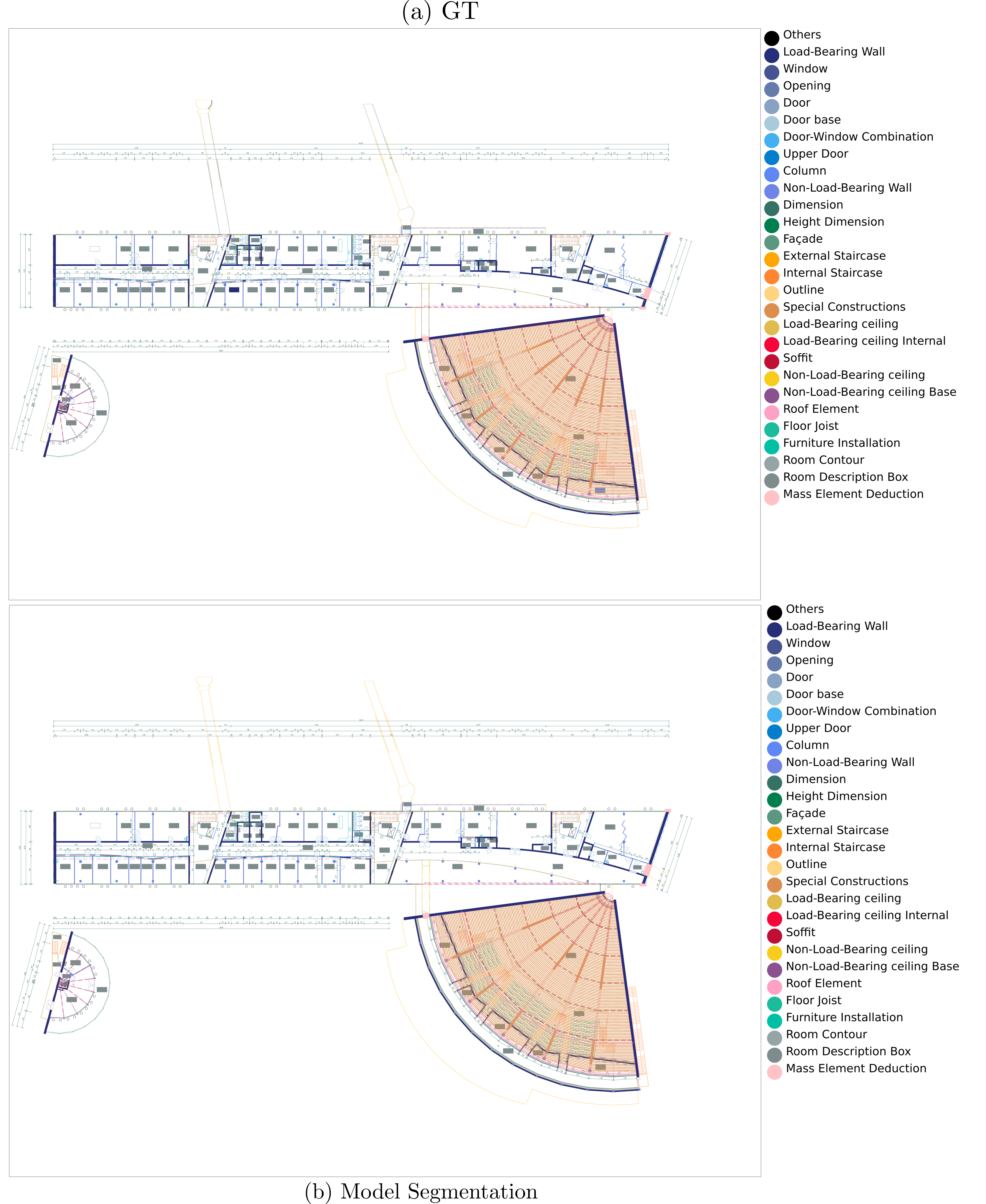}
\caption{Segmentation results comparison between ground truth and predicted labels on the TUM dataset}
\label{fig:Results_TUM1}
\end{figure}

\begin{table}[ht]
\small
\centering
\caption{Comparison of F1 and weighted F1 (wF1) scores on the TUM Dataset}
\label{tab:results_flooplanCAD}
\begin{tabular}{ccccccc}
\hline
Method & PanCADNet & CADTransformer & GAT-CADNet & PointT‡ & SymPoint  & VectorGraphNET\\
\hline
F1 & 80.6 & 82.2 & 85.0 & 83.2 & 86.8 & \textbf{79.4}\\
wF1 & 79.8 & 80.1 & 82.3 & 80.7 & 85.5 & \textbf{89.0}\\
\hline
\end{tabular}
\end{table}

We compare the results on the common test set with other neural networks that achieve line-level semantic segmentation on the technical drawings in FloorplanCAD and present the outcome in table \ref{tab:results_flooplanCAD}

\subsection{Results on TUM dataset}
The proprietary dataset was utilized for training and testing, with a random split of 62 drawings for training and 14 for testing.

The classification results, presented in table \ref{tab:classification_report_tum}, demonstrate that the proposed approach achieves an accuracy of 0.97, along with a macro average F1-score of 0.82 and a weighted average F1-score of 0.97. Notably, these results are reported at the most granular level of detail label.

\begin{figure}
\centering
\includegraphics[scale=0.72]{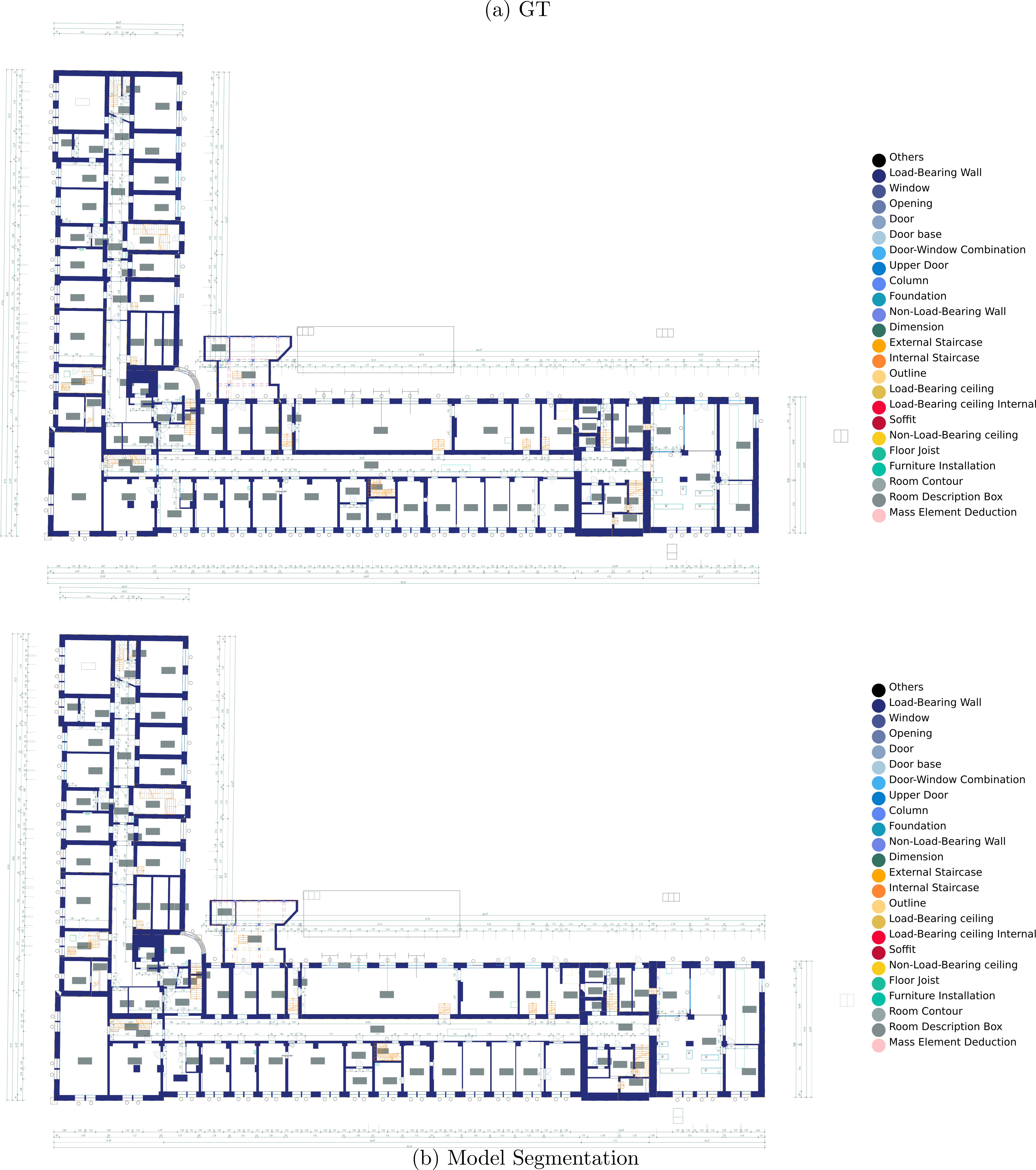}
\caption{Results on TUM Dataset}
\label{fig:Results_TUM2}
\end{figure}

\begin{figure}
\centering
\includegraphics[scale=0.79]{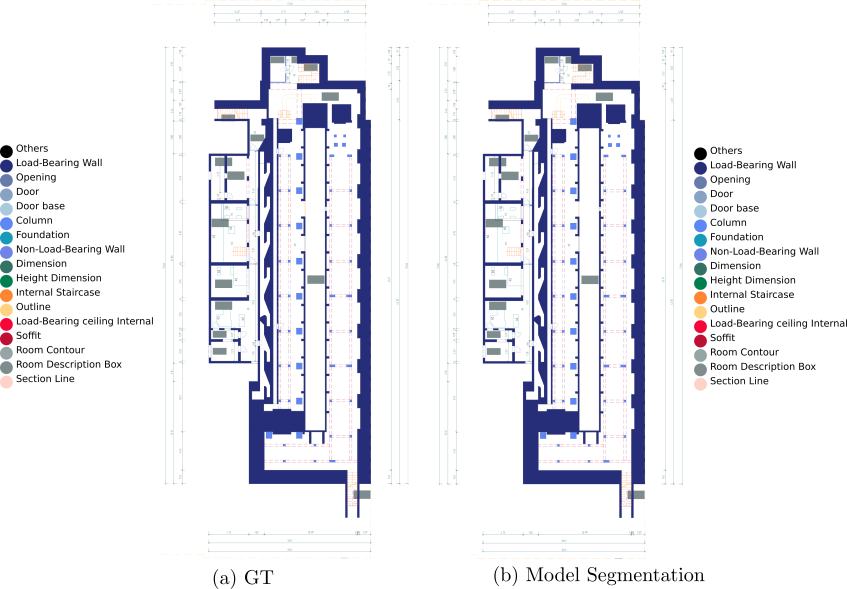}
\caption{Results on TUM Dataset}
\label{fig:Results_TUM3}
\end{figure}

A visual comparison of ground truth and model predictions is provided in Figure \ref{fig:Results_TUM1}, Figure \ref{fig:Results_TUM2}, and Figure \ref{fig:Results_TUM3}. Detailed classification report for each class is presented in Table \ref{tab:classification_report_tum}. In contrast to the other dataset analyzed, the proprietary dataset was used as input for the neural network model in its entirety, without slicing the information into patches. This approach allowed the model to maintain the overall design intact. Furthermore, whereas the previous dataset focused on symbol spotting, with most classes consisting of recurring symbols and other lines grouped into an undefined class, the proprietary dataset provided a separate and distinct class for most lines. This distinction enabled the segmentation to define better and clearer delimitations, making it a more suitable solution for line-level general-purpose predictions.

\begin{table}[ht]
\small
\centering

\caption{Classification Report for TUM Dataset}
\begin{tabular}{cccccc}
\hline
Class & Precision & Recall & F1-score & Support \\
\hline
Others	& 0.97 & 0.98 & 0.97 & 87689 \\		
Grid 	& 0.99 & 0.99 & 0.99 & 255879 \\		
Load-Bearing Wall 	& 0.96 & 0.95 & 0.95 & 55405 \\		
Window 	& 0.97 & 0.97 & 0.97 & 13721 \\		
Opening 	& 0.91 & 0.86 & 0.88 & 2020 \\		
Door 	& 0.91 & 0.91 & 0.91 & 5928 \\		
Door base 	& 0.93 & 0.88 & 0.91 & 4479 \\		
Door-Window Combination 	& 0.87 & 0.92 & 0.90 & 15570 \\	
Upper Door 	& 0.88 & 0.86 & 0.87 & 3028 \\		
Column 	& 0.95 & 0.98 & 0.96 & 9251 \\		
Foundation 	& 1.00 & 0.96 & 0.97 & 19215 \\		
Non-Load-Bearing Wall 	& 0.96 & 0.96 & 0.96 & 21071 \\		
Partition Wall 	& 0.98 & 0.92 & 0.95 & 212 \\		
Dimension 	& 0.99 & 0.98 & 0.98 & 52288 \\		
Height Dimension 	& 0.82 & 0.77 & 0.79 & 150 \\		
Façade 	& 0.93 & 0.87 & 0.90 & 1685 \\		
External Staircase 	& 0.84 & 0.87 & 0.86 & 1289 \\		
Internal Staircase	& 0.80 & 0.87 & 0.83 & 2829 \\		
Outline 	& 0.66 & 0.59 & 0.62 & 2106 \\		
Hidden Outline 	& 1.00 & 0.98 & 0.99 & 431 \\		
Special Constructions	& 0.91 & 0.87 & 0.89 & 3843 \\		
Load-Bearing ceiling 	& 0.78 & 0.85 & 0.81 & 3578 \\		
Load-Bearing ceiling 2D 	& 0.56 & 0.67 & 0.61 & 103 \\		
Soffit 	& 0.85 & 0.91 & 0.88 & 5448 \\		
Non-Load-Bearing ceiling 	& 0.79 & 0.84 & 0.82 & 900 \\		
Roof 	& 0.99 & 0.98 & 0.98 & 172 \\		
Roof Element 	& 0.89 & 0.99 & 0.93 & 813 \\		
Roof Construction 	& 0.00 & 0.00 & 0.00 &   0 \\		
Existing Building Hatch 	& 0.00 & 0.00 & 0.00 &   0 \\		
Building Connection 	& 0.98 & 0.99 & 0.98 & 1213 \\		
Roof Joist 	& 0.00 & 0.00 & 0.00 &   9 \\		
Floor Joist 	& 0.49 & 0.62 & 0.55 &  29 \\		
Furniture Installation 	& 1.00 & 0.96 & 0.98 & 15895 \\		
Special Installation 	& 1.00 & 1.00 & 1.00 & 1880 \\		
Sanitary Installation 	& 0.96 & 0.93 & 0.95 & 1037 \\		
Room Contour 	& 0.87 & 0.93 & 0.90 & 9807 \\		
Component Table 	& 0.91 & 0.97 & 0.94 & 1064 \\		
Mass Element Deduction 	& 0.90 & 0.70 & 0.79 & 134 \\		
Section Line 	& 0.36 & 0.84 & 0.50 &  57 \\		
Overview Details 	& 0.98 & 0.99 & 0.98 & 1920 \\		

\hline
accuracy &  &  & 0.97 & 602148 \\
macro avg & 0.81 & 0.83 & 0.82 & 602148 \\
weighted avg & 0.97 & 0.97 & 0.97 & 602148 \\
\hline
\end{tabular}
\label{tab:classification_report_tum}
\end{table}

\begin{table}[ht]
\small
\centering
\caption{Classification Report for FloorplanCAD dataset, Precision, Recall, F1-score refers to the training with hierarchical labels, Recall1, Precision1, F1\_score1 refer to the training without hierarchical labels.}
\begin{tabular}{ccccccccc}
\hline
Object & Class & Precision & Recall & F1-score & Support & Recall1 & Precision1 & F1\_score1 \\
\hline
single door & 0 & 0.85 & 0.9 & 0.87 & 724544 & 0.8 & 0.74 & 0.77 \\
double door & 1 & 0.93 & 0.88 & 0.9 & 1003280 & 0.89 & 0.81 & 0.85 \\
sliding door & 2 & 0.91 & 0.84 & 0.87 & 14238 & 0.92 & 0.86 & 0.88 \\
folding door & 3 & 0.86 & 0.81 & 0.83 & 60730 & 0.97 & 0.45 & 0.62 \\
revolving door & 4 & 0.89 & 0.88 & 0.89 & 227157 & 0 & 0 & 0.00 \\
rolling door & 5 & 0.92 & 0.91 & 0.91 & 56385 & 0 & 0 & 0.00 \\
window & 6 & 0.74 & 0.8 & 0.77 & 239842 & 0.7 & 0.61 & 0.65 \\
bay window & 7 & 0.92 & 0.91 & 0.91 & 944828 & 0.74 & 0.04 & 0.08 \\
blind window & 8 & 0.68 & 0.66 & 0.67 & 2909 & 0.78 & 0.61 & 0.69 \\
opening symbol & 9 & 0.32 & 0.13 & 0.19 & 1269 & 0.43 & 0.07 & 0.12 \\
sofa & 10 & 0.82 & 0.75 & 0.78 & 3695 & 0.71 & 0.66 & 0.68 \\
bed & 11 & 0.91 & 0.9 & 0.91 & 30988 & 0.88 & 0.91 & 0.90 \\
chair & 12 & 0.71 & 0.72 & 0.71 & 19922 & 0.78 & 0.57 & 0.66 \\
table & 13 & 0.8 & 0.52 & 0.63 & 9259 & 0.81 & 0.44 & 0.57 \\
TV cabinet & 14 & 0.88 & 0.85 & 0.86 & 2772 & 0.81 & 0.8 & 0.80 \\
Wardrobe & 15 & 0.94 & 0.9 & 0.92 & 14455 & 0.9 & 0.9 & 0.90 \\
cabinet & 16 & 0.84 & 0.69 & 0.76 & 10678 & 0.81 & 0.67 & 0.73 \\
gas stove & 17 & 0.96 & 0.89 & 0.93 & 11459 & 0.96 & 0.83 & 0.89 \\
sink & 18 & 0.89 & 0.84 & 0.86 & 50549 & 0.87 & 0.81 & 0.84 \\
refrigerator & 19 & 0.91 & 0.73 & 0.81 & 3679 & 0.81 & 0.67 & 0.73 \\
air-conditioner & 20 & 0.75 & 0.61 & 0.67 & 3737 & 0.71 & 0.59 & 0.64 \\
bath & 21 & 0.69 & 0.53 & 0.6 & 8639 & 0.62 & 0.4 & 0.49 \\
bathtub & 22 & 0.85 & 0.54 & 0.66 & 5065 & 0.65 & 0.63 & 0.64 \\
washing machine & 23 & 0.93 & 0.86 & 0.89 & 9396 & 0.9 & 0.82 & 0.86 \\
urinal & 24 & 0.96 & 0.94 & 0.95 & 16980 & 0.95 & 0.93 & 0.94 \\
squat toilet & 25 & 0.87 & 0.93 & 0.89 & 10852 & 0.81 & 0.93 & 0.86 \\
toilet & 26 & 0.88 & 0.93 & 0.9 & 37393 & 0.87 & 0.89 & 0.88 \\
stairs & 27 & 0.88 & 0.78 & 0.83 & 27750 & 0.88 & 0.75 & 0.81 \\
elevator & 28 & 0.86 & 0.87 & 0.86 & 13753 & 0.82 & 0.85 & 0.84 \\
escalator & 29 & 0.69 & 0.53 & 0.6 & 1710 & 0.57 & 0.55 & 0.56 \\
row chairs & 30 & 0.93 & 0.93 & 0.93 & 54675 & 0.91 & 0.96 & 0.93 \\
parking spot & 31 & 0.92 & 0.87 & 0.9 & 30029 & 0.93 & 0.86 & 0.90 \\
wall & 32 & 0.73 & 0.87 & 0.79 & 148085 & 0.7 & 0.82 & 0.76 \\
curtain wall & 33 & 0.72 & 0.44 & 0.55 & 24595 & 0.64 & 0.41 & 0.50 \\
railing & 34 & 0.45 & 0.37 & 0.41 & 4292 & 0.39 & 0.37 & 0.38 \\
other & 35 & 0.91 & 0.92 & 0.92 & 943546 & 0.9 & 0.91 & 0.90 \\
accuracy &  & \textbf{0.89} & & & 4773135 &  \textbf{0.86}\\
macro avg &  & \textbf{0.82} & \textbf{0.76} & \textbf{0.79} & 4773135 & \textbf{0.74} & \textbf{0.64} & \textbf{0.67}\\
weighted avg &  & \textbf{0.89} & \textbf{0.89} & \textbf{0.88} & 4773135 & \textbf{0.86} & \textbf{0.86} & \textbf{0.86}\\
\hline
\end{tabular}
\label{table:combarison_hierarchical}
\end{table}

\newpage
\section{Discussion} \label{sec:discussion}
We evaluate the performance of our segmentation approach on the FloorplanCAD dataset, comparing it to other state-of-the-art neural networks. While VectorGraphNET achieves a competitive F1 score of 79.4, slightly lower than the top-performing SymPoint method (86.8), it significantly outperforms all other techniques, including SymPoint, in terms of weighted F1 (wF1) with a score of 89.0. Notably, the wF1 metric is this task's most critical evaluation criterion, as it accounts for class imbalance and accurately represents the model's performance. This superior performance in wF1 highlights our method's effectiveness in addressing class imbalance and demonstrates its robustness in handling complex architectural drawings.

Our neural network architecture is lightweight, with only 1.3 million weights, compared to existing implementations such as PanCADNet (>42M), CADTransformer (>65M), PointT‡Cluster (31M), and SymPoint (35M), which all feature neural networks with over 30 million weights. This reduction in complexity is mainly due to our novel approach of precomputing an extensive graph, which, although computationally expensive upfront, enables efficient processing on a lighter GPU. By performing this graph computation on the CPU, we can leverage parallelization at a lower cost than parallelizing the neural network computation on multiple GPUs. Furthermore, this precomputation step allows us to operate on a simplified representation of the data, assuming that the post-processing of every vector file is correct and all vectors are in plain representation. As a result, our approach can be trained on a single NVIDIA RTX 8000, featuring 48 GB of GDDR6 memory and 672 GB/s of memory bandwidth, in contrast to SymPoint, which requires an impressive computational infrastructure of 8 NVIDIA A100 GPUs with a combined 384 GB of HBM2 memory and 12,440 GB/s of memory bandwidth. This highlights the efficiency and scalability of our proposed architecture. Our model demonstrates efficient training, requiring only 1 minute per epoch on the NVIDIA RTX 8000 GPU, and achieves a wF1 score of 0.85 after 50 epochs, with peak performance reached at 500 epochs.

Notably, our approach differs from existing methods, relying solely on geometrical vector information without considering CAD primitives or layers. This design choice makes our approach more generic and widely applicable to any drawing in vector format, regardless of its origin or structure. In contrast, methods that rely on CAD-specific information are inherently limited to CAD-specific applications, which restricts their usability. By avoiding CAD-specific dependencies, our approach is more robust to variations in CAD software, file formats, and user workflows and can adapt more easily to changes in the field.

\section{Conclusion} \label{sec:conclusion}
In this study, we presented a novel approach to semantic segmentation of architectural drawings using a hierarchical multi-label definition. Our method, VectorGraphNET, achieves state-of-the-art performance on the FloorplanCAD dataset, outperforming existing methods in weighted F1 score. Notably, our approach is remarkably lightweight, with only 1.3 million weights, making it more efficient and scalable than existing implementations.

Our results demonstrate the effectiveness of our approach in addressing class imbalance and handling complex architectural drawings. Using a hierarchical multi-label definition allows our model to capture the underlying patterns in the data, leading to improved classification performance. Furthermore, our approach is more generic and widely applicable to any drawing in vector format, regardless of its origin or structure.

The efficiency and scalability of our proposed architecture make it an attractive solution for real-world applications. 

In conclusion, our work presents a significant contribution to the field of semantic segmentation of architectural drawings. Our approach offers a more efficient, scalable, and widely applicable solution for real-world applications, and we believe that it has the potential to be used in a variety of fields, including architecture, engineering, and construction.

\section{Data Availability} \label{sec:availability}
The FloorPlanCAD dataset is publicly available for download on the author's website \cite{fan2021floorplancadlargescalecaddrawing}. The TUM dataset is proprietary and, due to restrictions, is not currently available for sharing. 
\section{Acknowledgement} \label{sec:acknowledgment}
The presented research has received funding from the Bayerisches Verbundforschungsprogramm (BayVFP) des Freistaates Bayern - Förderlinie "Digitalisierung", grant number DIK0336/01.
The authors would like to thank TUM Central Services for providing the CAD drawings used in this project.


\begin{thebibliography}{10}

\bibitem{borrmann2018building}
Andr{\'e} Borrmann, Markus K{\"o}nig, Christian Koch, and Jakob Beetz.
\newblock {\em Building information modeling: Why? what? how?}
\newblock Springer, 2018.

\bibitem{moreno2019new}
Carlos~Francisco Moreno-Garc{\'\i}a, Eyad Elyan, and Chrisina Jayne.
\newblock New trends on digitisation of complex engineering drawings.
\newblock {\em Neural computing and applications}, 31:1695--1712, 2019.

\bibitem{9080390}
Waldir Lazaro-Aleman, Fernando Manrique-Galdos, Cesar Ramirez-Valdivia, Carlos Raymundo-Ibañez, and Javier~M. Moguerza.
\newblock Digital transformation model for the reduction of time taken for document management with a technology adoption approach for construction smes.
\newblock In {\em 2020 9th International Conference on Industrial Technology and Management (ICITM)}, pages 1--5, 2020.

\bibitem{RASMUSSEN2019102956}
Mads~Holten Rasmussen, Maxime Lefrançois, Pieter Pauwels, Christian~Anker Hviid, and Jan Karlshøj.
\newblock Managing interrelated project information in aec knowledge graphs.
\newblock {\em Automation in Construction}, 108:102956, 2019.

\bibitem{tariq2023study}
Junaid Tariq and S~Shujaa~Safdar Gardezi.
\newblock Study the delays and conflicts for construction projects and their mutual relationship: A review.
\newblock {\em Ain Shams Engineering Journal}, 14(1):101815, 2023.

\bibitem{carrara2024employing}
Andrea Carrara, Stavros Nousias, and A~Borrmann.
\newblock Employing graph neural networks for construction drawing content recognition.
\newblock In {\em i3CE 2024: 2024 ASCE International Conference on Computing in Civil Engineering}, 2024.

\bibitem{Mac2010AST}
S{\'e}bastien Mac{\'e}, Herv{\'e} Locteau, Ernest Valveny, and Salvatore Tabbone.
\newblock A system to detect rooms in architectural floor plan images.
\newblock In {\em International Workshop on Document Analysis Systems}, 2010.

\bibitem{Ahmed2011ImprovedAA}
Sheraz Ahmed, Marcus Liwicki, Markus Weber, and Andreas~R. Dengel.
\newblock Improved automatic analysis of architectural floor plans.
\newblock {\em 2011 International Conference on Document Analysis and Recognition}, pages 864--869, 2011.

\bibitem{kalervo2019cubicasa5kdatasetimprovedmultitask}
Ahti Kalervo, Juha Ylioinas, Markus Häikiö, Antti Karhu, and Juho Kannala.
\newblock Cubicasa5k: A dataset and an improved multi-task model for floorplan image analysis, 2019.

\bibitem{8237503}
Chen Liu, Jiajun Wu, Pushmeet Kohli, and Yasutaka Furukawa.
\newblock Raster-to-vector: Revisiting floorplan transformation.
\newblock In {\em 2017 IEEE International Conference on Computer Vision (ICCV)}, pages 2214--2222, 2017.

\bibitem{7986875}
Samuel Dodge, Jiu Xu, and Björn Stenger.
\newblock Parsing floor plan images.
\newblock In {\em 2017 Fifteenth IAPR International Conference on Machine Vision Applications (MVA)}, pages 358--361, 2017.

\bibitem{10.1145/3441250.3441265}
Wei Wang, Shuai Dong, Kun Zou, and Wen-sheng LI.
\newblock Room classification in floor plan recognition.
\newblock In {\em Proceedings of the 4th International Conference on Advances in Image Processing}, ICAIP '20, page 48–54, New York, NY, USA, 2021. Association for Computing Machinery.

\bibitem{9577792}
Xiaolei Lv, Shengchu Zhao, Xinyang Yu, and Binqiang Zhao.
\newblock Residential floor plan recognition and reconstruction.
\newblock In {\em 2021 IEEE/CVF Conference on Computer Vision and Pattern Recognition (CVPR)}, pages 16712--16721, 2021.

\bibitem{9009528}
Zhiliang Zeng, Xianzhi Li, Ying~Kin Yu, and Chi-Wing Fu.
\newblock Deep floor plan recognition using a multi-task network with room-boundary-guided attention.
\newblock In {\em 2019 IEEE/CVF International Conference on Computer Vision (ICCV)}, pages 9095--9103, 2019.

\bibitem{10.1007/978-981-15-6648-6_2}
Ilya~Y. Surikov, Mikhail~A. Nakhatovich, Sergey~Y. Belyaev, and Daniil~A. Savchuk.
\newblock Floor plan recognition and vectorization using combination unet, faster-rcnn, statistical component analysis and ramer-douglas-peucker.
\newblock In Nirbhay Chaubey, Satyen Parikh, and Kiran Amin, editors, {\em Computing Science, Communication and Security}, pages 16--28, Singapore, 2020. Springer Singapore.

\bibitem{zhang2020directionawarelearnableadditivekernels}
Yuli Zhang, Yeyang He, Shaowen Zhu, and Xinhan Di.
\newblock The direction-aware, learnable, additive kernels and the adversarial network for deep floor plan recognition, 2020.

\bibitem{info12050206}
Shuai Dong, Wei Wang, Wensheng Li, and Kun Zou.
\newblock Vectorization of floor plans based on edgegan.
\newblock {\em Information}, 12(5), 2021.

\bibitem{10.1145/3210499.3210528}
Toshihiko Yamasaki, Jin Zhang, and Yuki Takada.
\newblock Apartment structure estimation using fully convolutional networks and graph model.
\newblock In {\em Proceedings of the 2018 ACM Workshop on Multimedia for Real Estate Tech}, RETech'18, page 1–6, New York, NY, USA, 2018. Association for Computing Machinery.

\bibitem{10.1145/3638584.3638636}
Wenming Wu.
\newblock Architectural floorplan recognition via iterative semantic segmentation networks.
\newblock In {\em Proceedings of the 2023 7th International Conference on Computer Science and Artificial Intelligence}, CSAI '23, page 282–287, New York, NY, USA, 2024. Association for Computing Machinery.

\bibitem{9506514}
Christoffer~P. Simonsen, Frederik~M. Thiesson, Mark~P. Philipsen, and Thomas~B. Moeslund.
\newblock Generalizing floor plans using graph neural networks.
\newblock In {\em 2021 IEEE International Conference on Image Processing (ICIP)}, pages 654--658, 2021.

\bibitem{7410526}
Ross Girshick.
\newblock Fast r-cnn.
\newblock In {\em 2015 IEEE International Conference on Computer Vision (ICCV)}, pages 1440--1448, 2015.

\bibitem{fan2021floorplancadlargescalecaddrawing}
Zhiwen Fan, Lingjie Zhu, Honghua Li, Xiaohao Chen, Siyu Zhu, and Ping Tan.
\newblock Floorplancad: A large-scale cad drawing dataset for panoptic symbol spotting, 2021.

\bibitem{zheng2022gatcadnetgraphattentionnetwork}
Zhaohua Zheng, Jianfang Li, Lingjie Zhu, Honghua Li, Frank Petzold, and Ping Tan.
\newblock Gat-cadnet: Graph attention network for panoptic symbol spotting in cad drawings, 2022.

\bibitem{9879621}
Zhiwen Fan, Tianlong Chen, Peihao Wang, and Zhangyang Wang.
\newblock Cadtransformer: Panoptic symbol spotting transformer for cad drawings.
\newblock In {\em 2022 IEEE/CVF Conference on Computer Vision and Pattern Recognition (CVPR)}, pages 10976--10986, 2022.

\bibitem{liu2024symbolpointspanopticsymbol}
Wenlong Liu, Tianyu Yang, Yuhan Wang, Qizhi Yu, and Lei Zhang.
\newblock Symbol as points: Panoptic symbol spotting via point-based representation, 2024.

\bibitem{zhao2021pointtransformer}
Hengshuang Zhao, Li~Jiang, Jiaya Jia, Philip Torr, and Vladlen Koltun.
\newblock Point transformer, 2021.

\bibitem{cheng2022maskedattentionmasktransformeruniversal}
Bowen Cheng, Ishan Misra, Alexander~G. Schwing, Alexander Kirillov, and Rohit Girdhar.
\newblock Masked-attention mask transformer for universal image segmentation, 2022.

\bibitem{iso32000}
Document management -- portable document format -- part 1: Pdf 1.7.
\newblock \url{https://www.iso.org/standard/51502.html}, 2008.
\newblock [Accessed 26-09-2024].

\bibitem{w3ScalableVector}
{S}calable {V}ector {G}raphics ({S}{V}{G}) 2.0 {S}pecification --- dev.w3.org.
\newblock \url{https://dev.w3.org/SVG/profiles/2.0/publish/single-page.html}.
\newblock [Accessed 26-09-2024].

\bibitem{carlier2020deepsvghierarchicalgenerativenetwork}
Alexandre Carlier, Martin Danelljan, Alexandre Alahi, and Radu Timofte.
\newblock Deepsvg: A hierarchical generative network for vector graphics animation, 2020.

\bibitem{brody2021attentive}
Shaked Brody, Uri Alon, and Eran Yahav.
\newblock How attentive are graph attention networks?, 2021.

\bibitem{bayern}
VHF Bayern.
\newblock Vi.4.2.h layerstruktur.
\newblock \url{https://www.stmb.bayern.de/assets/stmi/buw/bauthemen/vergabeundvertragswesen/vhf/vi-4-2_layerliste_2018.pdf}, 2018.
\newblock [Accessed 27-09-2024].

\end{thebibliography}

\section{Notation}
\label{app:notation}
\emph{The following symbols are used in this paper:}
\nopagebreak
\par
\begin{tabular}{r  @{\hspace{1em}=\hspace{1em}}  l}
CAD                    & Computer-Aided Design; \\
AEC                    & Architecture, Engineering, and Construction; \\
PLM                    & product lifecycle management; \\
PDF                    & Portable Document Format; \\
GNN                    & Graph Neural Network; \\
SVG                    & Scalable Vector Graphics; \\
KNN                    & K-nearest neighbor; \\
wF1                    & weighted F1 score; \\
GPU                    & Graphics Processing Unit; \\
HBM2                   & High-Bandwidth Memory 2; \\
GDDR6                  & Graphics Double Data Rate 6; \\
F1                     & F1 score; \\
GB                     & Giga Byte. \\
\end{tabular}

%
%

%

\end{document}